\documentclass[lettersize,journal]{IEEEtran}

\usepackage{amsmath,amsfonts}
\usepackage{bm}
\usepackage{algorithm}
\usepackage{algpseudocode}
\usepackage{array}
\usepackage[font=normalsize,labelfont=sf,textfont=sf]{subcaption}
\usepackage{textcomp}
\usepackage{stfloats}
\usepackage{url}
\usepackage{verbatim}
\usepackage{graphicx}
\def\BibTeX{{\rm B\kern-.05em{\sc i\kern-.025em b}\kern-.08em
    T\kern-.1667em\lower.7ex\hbox{E}\kern-.125emX}}
\usepackage{balance}

\usepackage{epstopdf}
\usepackage{multirow}
\usepackage{hyperref}
\usepackage{tabularx}
\usepackage{makecell}
\usepackage{booktabs}
\usepackage{placeins}
\usepackage[T1]{fontenc}

\begin{document}
\title{HiSE: A Lightweight Hierarchical Semantic Explainer for Heterogeneous Graph Neural Networks}
\author{
\textbf{Zongrui Li}\textsuperscript{1},
\textbf{Yuhang Zhao}\textsuperscript{1},
\textbf{Ying Zhao}\textsuperscript{1},  \\
\textbf{Yuanzhao Guo}\textsuperscript{1},
\textbf{Qiang Huang}\textsuperscript{2},
\textbf{Yuan Tian}\textsuperscript{1} \\
\vspace{0.5em}
\textsuperscript{1}School of Artificial Intelligence, Jilin University \\
\textsuperscript{2}Mohamed bin Zayed University of Artificial Intelligence
}


\maketitle

\begin{abstract}

Heterogeneous graph neural networks (HGNNs) have demonstrated remarkable performance in modeling complex relational data, however their interpretability in high-stakes applications remains a critical challenge. Existing explanation methods suffer from two major limitations: on the one hand, the generated explanations fail to reflect the inherent semantic hierarchy of HGNNs, resulting in a lack of fidelity to the model's internal decision-making mechanism; on the other hand, feature explanations often rely on complex search or perturbation mechanisms, leading to excessive computational complexity and poor efficiency. To address these issues, we propose HiSE, a lightweight feature-oriented interpretable model for HGNNs. HiSE achieves semantically aware feature explanations through hierarchical semantic modeling: at the \textit{semantic level}, local surrogate models based on the Least Absolute Shrinkage and Selection Operator (LASSO) are employed to learn sparse feature representations under each semantic view; at the \textit{cross-semantic level}, the contributions of different semantic views are adaptively characterized via KL divergence to produce a unified explanation. Extensive experiments demonstrate that HiSE outperforms existing methods in terms of fidelity, robustness, and cross-semantic explanation capability, while its lightweight framework incurs low computational overhead, enabling efficient application to large-scale, complex real-world heterogeneous graphs.

\end{abstract}


\section{Introduction}

\begin{figure}[htbp]
\centering
{
    \begin{minipage}[b]{.98\linewidth}
    \centering
    \includegraphics[width=\linewidth]{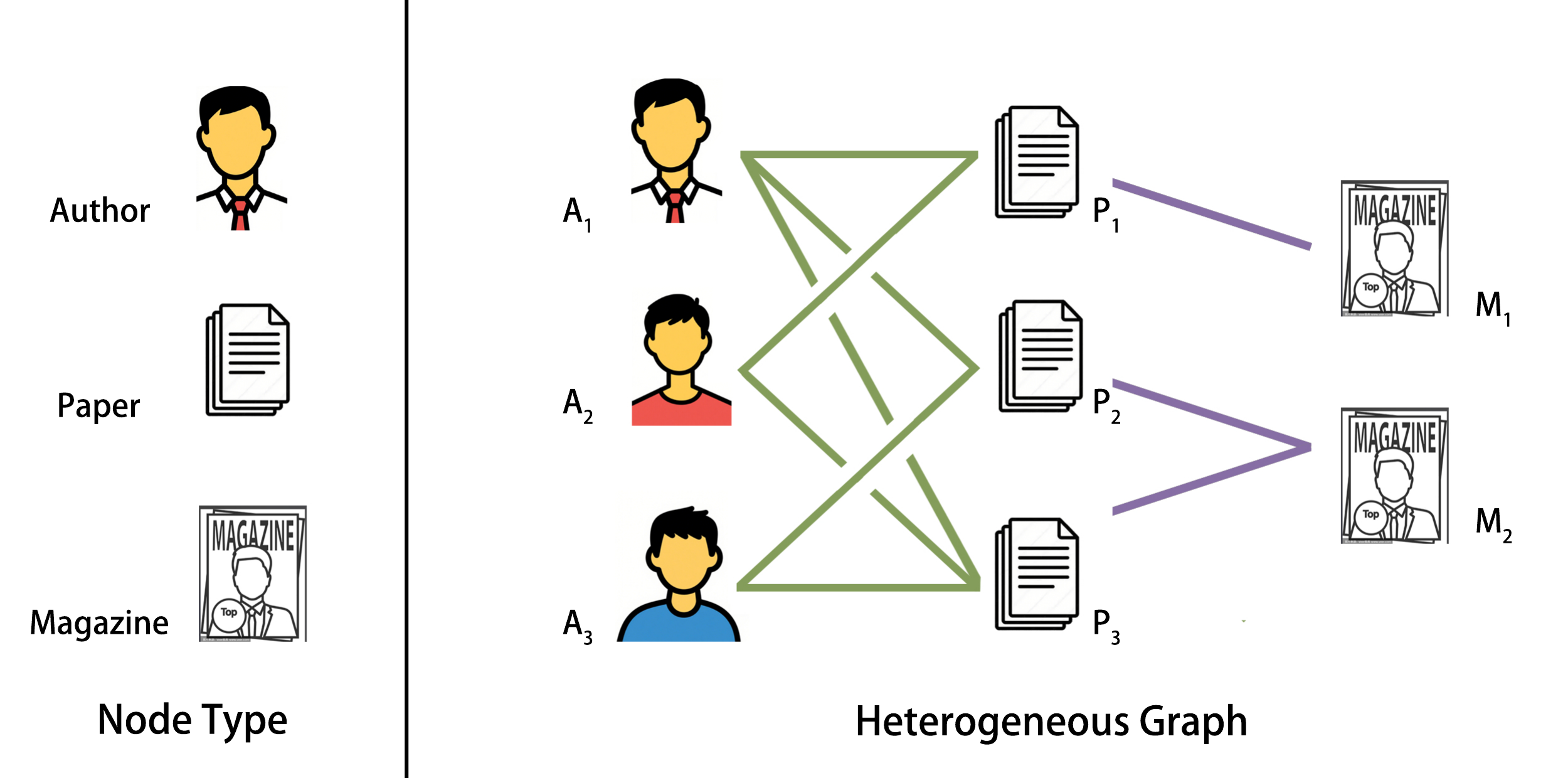}
    \end{minipage}
}
\caption{A schematic representation of the heterogeneous graph. The graph consists of three types of nodes: authors, papers, and magazines. Edges connect authors to papers and papers to magazines, illustrating the heterogeneous structure where multiple node types and relation types coexist.}
\label{Schematic1}
\end{figure}

\IEEEPARstart{G}{raph} neural networks (GNNs)~\cite{TheGNN, 9046288, ZHOU202057} have emerged as powerful frameworks for learning representations of graph-structured data by propagating and aggregating information among nodes via message-passing mechanisms~\cite{gilmer2017neuralmessagepassingquantum, Battaglia2018RelationalIB}, giving rise to representative models such as GAT~\cite{veličković2018graphattentionnetworks} and GraphMAE~\cite{10.1145/3534678.3539321}. However, conventional GNNs assume node and edge homogeneity, limiting them to homogeneous scenarios with a single node and edge type~\cite{wang2020surveyheterogeneousgraphembedding}. In practice, complex real-world systems typically consist of diverse entity and relation types, necessitating their representation as heterogeneous graphs~\cite{7536145}, where nodes and edges of different types carry rich and diverse semantic information that is essential for accurate modeling. For instance, the academic network illustrated in Fig.~\ref{Schematic1} contains three types of nodes—authors (A), papers (P), and magazines (M)—and multiple edge types. Representing such data as a homogeneous graph would flatten these types of distinctions, making it impossible to distinguish semantically different interactions such as coauthorship and authorship. By preserving node and edge type information, heterogeneous graphs enable the explicit modeling of cross-type semantic structures, thereby capturing complex real-world relationships more faithfully. To effectively learn node representations on such heterogeneous graphs, heterogeneous graph neural networks (HGNNs)~\cite{PathSim, 10.1145/3366423.3380027, LIU2024127274, guan2025heterogeneous} have been proposed. The core idea of HGNNs lies in hierarchical semantic modeling: capturing individual semantics arising from diverse relation types at the semantic level and progressively integrating them into richer cross-semantic associations at the cross-semantic level. On the basis of the semantic modeling strategy, existing HGNNs can be broadly categorized into \textit{meta-path-based} and \textit{relation-based} approaches~\cite{article}. The former explicitly extracts semantic subgraphs via predefined meta-paths and employs attention mechanisms to assign weights to different semantics, with representative works including HAN~\cite{wang2021heterogeneousgraphattentionnetwork}, which adopts a dual-level attention mechanism to integrate multi-path semantic information. The latter is employed to directly model type-aware message passing over heterogeneous edges and learn intersemantic hierarchical relationships in an implicit manner, as exemplified by HetSANN~\cite{hong2019attentionbasedgraphneuralnetwork}, which leverages a heterogeneous self-attention network to capture semantic dependencies among nodes. Compared with conventional GNNs, HGNNs can hierarchically model semantic heterogeneity and better capture cross-type relations, yielding more accurate and robust representations~\cite{HGNN, article, 10008205}, and have demonstrated significant advantages in domains such as recommender systems~\cite{9925059, 10.1145/3568395, 10132398}, knowledge graph completion~\cite{11204706}, and drug discovery~\cite{10059171}.

In both homogeneous and heterogeneous graphs, GNNs must balance high predictive accuracy with interpretability in real-world applications~\cite{BARREDOARRIETA202082,yuan2022explainabilitygraphneuralnetworks, li2023surveyexplainablegraphneural}. For example, in health care, GNNs can integrate multimodal information—such as electronic health records, medical knowledge graphs, and clinical trial data—to support disease prediction, where explanations are essential for clinicians to trust model outputs and make informed decisions~\cite{10115472, cui2022interpretablegraphneuralnetworks, OSSBOLL2024104616}. Furthermore, in numerous application scenarios such as financial fraud detection~\cite{9609099, MOTIE2024122156,10478631}, interpretability remains a key factor in ensuring model trustworthiness, transparency, and practical deployability~\cite{10582518}.
Substantial progress has been made in improving the interpretability of GNNs. For instance, GNNExplainer~\cite{ying2019gnnexplainergeneratingexplanationsgraph} and GraphLIME~\cite{huang2020graphlimelocalinterpretablemodel} identify critical subgraphs and features by learning mutual-information-driven masks and constructing local surrogate models, respectively, thereby providing explanations for node predictions. 
However, despite their powerful semantic modeling capabilities, explaining HGNNs remains difficult because of the intricate interplay of multiple node and edge types in heterogeneous graphs~\cite{7536145}.
Moreover, existing methods focus predominantly on structural explanations, while the analysis of node feature contributions remains largely underexplored. For example, xPath~\cite{xPath} identifies important structures by searching for critical paths that significantly influence model predictions, providing structural-level explanations for HGNNs; Tail~\cite{tail} characterizes the evolution of research interests by modeling key interaction paths in temporal heterogeneous graphs, thereby offering interpretable analysis; and PaGE-Link~\cite{pagelink}, which focuses on heterogeneous link prediction tasks, identifies the most influential semantic paths on the basis of path-level representation learning.
On the other hand, a few methods have begun to address feature explanations. For instance, HGExplainer~\cite{HGExplainer} extends GNNExplainer~\cite{ying2019gnnexplainergeneratingexplanationsgraph} to heterogeneous graphs by learning masks to identify the most predictive substructures and the contributing node features; HENCE-X~\cite{HENCEX} identifies important features by searching for critical feature subsets that preserve model prediction consistency. However, these methods all rely on perturbation-based explanation strategies—that is, evaluating importance by masking features and observing changes in model predictions—which share the following common limitations. First, these strategies do not capture the semantic hierarchy, making accurate characterization of how HGNNs make decisions across different semantic levels difficult. Second, feature perturbations can disrupt the original data distribution, compromising the stability and reliability of explanations in high-dimensional or noisy scenarios. Furthermore, computational efficiency remains a shared challenge among these methods, with HENCE-X relying on a greedy search strategy that requires repeated evaluation of candidate feature subsets, incurring particularly high computational overhead.

On the basis of the above analysis, our research motivation stems from the following observations: 
mainstream HGNN models rely on semantic modeling and cross-semantic information fusion to capture complex semantics in heterogeneous graphs, whereas existing explanation methods lack a hierarchical perspective and fail to effectively reflect the decision-making mechanisms of HGNNs.
Furthermore, existing explainers often rely on complex search or perturbation strategies, resulting in suboptimal computational efficiency. To address these issues, we propose \textbf{HiSE} (\textbf{Hi}erarchical \textbf{S}emantic \textbf{E}xplainer), a lightweight hierarchical semantic feature explanation framework for HGNNs that mirrors the model’s own hierarchical structure. At the \textit{semantic level}, HiSE LASSO-based local surrogate models is employed to learn sparse feature representations under each semantic view; at the \textit{cross-semantic level}, the contributions of different semantics are adaptively measured via KL divergence and fused to produce a unified explanation. HiSE not only aligns with the intrinsic decision-making mechanism of HGNNs to provide intuitive hierarchical explanations but also achieves high computational efficiency without relying on complex search or perturbation procedures.

The main contributions of this paper are summarized as follows:
\begin{itemize}
\item \textbf{We propose a new hierarchical semantic explanation paradigm for HGNNs.}
Starting from the intrinsic modeling mechanism of HGNNs, we reformulate the explanation problem as a hierarchical modeling process over multirelational semantic structures, presenting an explanation approach that is faithful to the model's decision-making mechanism. This paradigm simultaneously characterizes information contributions at both the semantic and cross-semantic levels, offering a new perspective for interpretability research on HGNNs.

\item \textbf{We design HiSE, a novel hierarchical semantic feature explanation framework.} 
HiSE constructs local surrogate models at the semantic level and comprehensively evaluates semantic importance at the cross-semantic level, thereby more accurately characterizing the internal decision-making mechanism of HGNNs.

\item \textbf{We conduct systematic experiments to validate the efficiency and effectiveness of the proposed method.}
Extensive experiments on multiple real-world heterogeneous graph models and datasets evaluate the proposed method from multiple perspectives, including fidelity, robustness, usability, and cross-semantic explanation capability. The experimental results demonstrate that HiSE consistently outperforms existing explanation methods across different experimental scenarios and is capable of stably identifying the most influential features for model predictions in complex feature spaces while incurring low computational overhead.
\end{itemize}

\section{Related Work}

This section provides the necessary background for our work. We first review the development of graph neural networks, from homogeneous architectures to heterogeneous extensions that model complex multityped relational data. We then survey existing interpretability methods for GNNs, covering both techniques designed for homogeneous graphs and recent efforts tailored to the heterogeneous setting.

\subsection{From Graph Neural Networks to Heterogeneous Architectures}

By propagating and aggregating information among nodes via message-passing mechanisms, graph neural networks (GNNs) have become the predominant approach for learning representations of graph-structured data. Representative works include GCN~\cite{kipf2017semisupervisedclassificationgraphconvolutional}, which leverages spectral graph convolutions for efficient neighbor feature aggregation; GAT~\cite{veličković2018graphattentionnetworks}, which introduces attention mechanisms to dynamically learn neighbor weights; and GraphMAE~\cite{10.1145/3534678.3539321}, which enhances graph representation capability through masked reconstruction tasks in a self-supervised manner. However, these methods all assume node and edge homogeneity, making them inadequate for handling heterogeneous graphs that contain multiple entity and relation types. To address this challenge, researchers have proposed heterogeneous graph neural networks (HGNNs), which can be broadly categorized into \textit{meta-path-based} and \textit{relation-based} approaches~\cite{article}. The former relies on predefined meta-paths to explicitly represent hierarchical semantic relations; for example, HAN~\cite{wang2021heterogeneousgraphattentionnetwork} fuses multi-path information through node-level and semantic-level hierarchical attention mechanisms; MAGNN~\cite{Fu_2020} further encodes intrameta-path structures to preserve fine-grained semantics; and HetGNN~\cite{chen2023srhetgnnsessionbasedrecommendationheterogeneousgraph} combines heterogeneous neighbor sampling with RNN-based aggregation strategies to learn node representations. The latter implicitly captures hierarchical semantic information in heterogeneous graphs from the perspective of relation types; for example, RGCN~\cite{schlichtkrull2017modelingrelationaldatagraph} introduces relation-specific weight matrices to implicitly model hierarchical differences among relations during message passing; HGT~\cite{10.1145/3366423.3380027} leverages transformer-based type-aware attention mechanisms to adaptively fuse node and relation information at a fine-grained level, and HetSANN~\cite{hong2019attentionbasedgraphneuralnetwork} models latent hierarchical semantics directly from relational structures through relation-aware self-attention aggregation, without requiring explicit meta-paths. In summary, these two categories of methods approach hierarchical semantic capture in heterogeneous graphs from the perspectives of explicit semantic modeling and implicit relation modeling, respectively, both advancing the capacity of HGNNs to encode hierarchical semantics in heterogeneous structures.

\subsection{Explainability of Graph Neural Networks} 

To improve the interpretability of GNNs, several explanation methods have been proposed. By optimizing masks that maximize mutual information with model predictions, GNNExplainer~\cite{ying2019gnnexplainergeneratingexplanationsgraph} identifies important subgraphs and node features. PGExplainer~\cite{luo2020parameterizedexplainergraphneural} adopts a probabilistic approach to learn global explanations across multiple instances. SubgraphX~\cite{yuan2021explainabilitygraphneuralnetworks} improves interpretability by searching for the most informative subgraphs through a Monte Carlo tree search. GraphLIME~\cite{huang2020graphlimelocalinterpretablemodel} extends LIME~\cite{ribeiro2016whyitrustyou} to graph structured data by learning nonlinear interpretable surrogate models. However, these methods are designed for homogeneous graphs and cannot directly handle the semantic heterogeneity and complex relational structures in HGNNs.
To address this gap, recent efforts have developed explanation methods tailored to HGNNs. One line of work approaches the problem from a structural perspective, characterizing model decision rationales through paths or subgraphs. For instance, xPath~\cite{xPath} identifies important structures by searching for critical paths that significantly influence model predictions, providing structural-level explanations for HGNNs; Tail~\cite{tail} models key interaction paths in temporal heterogeneous graphs to characterize the evolution of research interests; PaGE-Link~\cite{pagelink}, which focuses on link prediction tasks, identifies the most influential semantic paths through path-level representation learning; and HTGExplainer~\cite{HTGExplainer} further extends explanations to temporal heterogeneous graphs by generating explanatory subgraphs that preserve temporal dependencies and heterogeneity to capture dynamic semantics. Another line of work approaches the problem from a feature perspective, characterizing model decision rationales by selecting important features upon which model decisions rely. For example, HENCE-X~\cite{HENCEX} provides a model-agnostic explanation framework that uniformly characterizes topological and feature contributions through causality-guided conditional probabilities; HGExplainer~\cite{HGExplainer} combines mutual information maximization with meta-path-based sampling strategies to generate explanations from both structural and feature perspectives. Although the above methods improve HGNN interpretability from perspectives such as structural complexity and temporal dynamics, they do not explicitly align with the intrinsic semantic modeling mechanisms of HGNNs, making it difficult to characterize hierarchical semantic structures that are composed of multiple relation types and semantic paths. This limitation may lead to explanations that deviate from the model's true semantic reasoning process.

\section{Background and problem statement}

\begin{table}[h]
\centering
\caption{NOMENCLATURE}
\label{tab:nomenclature}
\begin{tabularx}{\columnwidth}{lX}
\hline
\textbf{Symbol} & \textbf{Description} \\
\hline
$n$ & Dimensionality of the feature space. \\
$m$ & The number of meta-paths. \\
$M$ & The HGNN to be explained. \\
$\Phi_k$ & Meta-path $k$. \\
$\mathcal{P}$ & Set of all meta-paths. \\
$\mathcal{G}$ & Complete heterogeneous graph. \\
$\mathcal{V}$ & Node set of heterogeneous graph. \\
$\mathcal{E}$ & Edge set of heterogeneous graph. \\
$\mathcal{N}$ & Set of all nodes belonging to the target node type. \\
$v_t$ & Target node to be explained.  \\
$X_i$ & Feature vector of node $i$. \\
$G_k$ & Single-semantic subgraph generated according to $\Phi_k$. \\
$V_k$ & Node set of $G_k$. \\
$E_k$ & Edge set of $G_k$. \\
$N_k$ & Set of sampled neighbors on $G_k$ from the target node. \\
$K$ & Maximum hop distance for neighborhood sampling. \\
$\mathbf{A}$ & Adjacency matrix.  \\

$\eta_{uv}$ & Weight assigned to edge $(u,v)$ by the explainer model $M$.  \\
$\mu_i^{(k)}$ & Normalized weight of node $v_i$ under meta-path $\Phi_k$.  \\
$\bm{s}_k$ & Semantic-level explanation vector for meta-path $\Phi_k$. \\
$L$ & Number of prediction classes.  \\
$\lambda$ & Regularization strength hyperparameter of LASSO. \\
$\bm{\hat{y}}_i$ & Classification probabilities of node $i$. \\
$\mathcal{D}_k$ & Local dataset obtained from model $M$ over nodes in $N_k$. \\
$\boldsymbol{w}$ & Vector of all semantic weights.  \\
$\mathbf{c}$ & Cross-semantic-level explanation vector. \\

\hline
\end{tabularx}
\end{table}

\subsection{Background}

Graph-structured data modeling has evolved significantly from homogeneous to heterogeneous representations, whose inherent structural diversity and semantic richness have given rise to heterogeneous graphs along with a range of modeling and analysis methods. This section introduces the fundamental concepts that underpin our work. We first formally define heterogeneous graphs and their structural properties, then introduce meta-paths as a key tool for representing semantic relationships in such graphs, and finally discuss HGNNs as a powerful learning framework for modeling these complex data structures.

\subsubsection{Heterogeneous Graph} A graph is considered \textit{heterogeneous} if it contains multiple types of nodes or edges; otherwise, it is considered \textit{homogeneous}. The academic network illustrated in Fig.~\ref{Schematic1} exemplifies a typical heterogeneous graph featuring three distinct node types: \textit{Papers} (P), characterized by data features such as publication content and keywords; \textit{Authors} (A), characterized by affiliation and research profile attributes; and \textit{Magazines} (M), representing publication venues with domain and prestige information. These nodes are interconnected through multiple edge types, including \textit{authorship} edges linking authors to their contributed papers and \textit{publication} edges linking papers to their published magazines. This complex structure enables the graph to preserve rich academic semantics that would be largely lost in a homogeneous representation, where all nodes and edges would be forced into a single type, thereby losing critical semantic distinctions.

Formally, a heterogeneous graph is defined as follows:
\[
\mathcal{G} = (\mathcal{V}, \mathcal{E})
\]
where $\mathcal{V}$ denotes the set of nodes and $\mathcal{E}$ denotes the set of edges. The structure also includes the following:
\begin{itemize}
    \item A node type mapping function: $\phi: \mathcal{V} \rightarrow \mathcal{Y}$
    \item An edge type mapping function: $\psi: \mathcal{E} \rightarrow \mathcal{R}$
\end{itemize}
subject to the following condition:
\[
|\mathcal{Y}| + |\mathcal{R}| > 2
\]
where $\mathcal{Y}$ and $\mathcal{R}$ denote the sets of node types and edge types, respectively.

\subsubsection{Meta-Path}

Meta-paths provide a powerful tool for representing the semantic relations of heterogeneous graphs and are widely employed in heterogeneous graph learning models~\cite{PathSim,Dong2017metapath2vecSR, wang2021heterogeneousgraphattentionnetwork}. While heterogeneous graphs offer powerful models for real-world systems with diverse entity and relation types, they also introduce the challenge of how to effectively model such data for complex reasoning tasks. A prerequisite for achieving this goal is integrating heterogeneous structural information and semantics composed of different nodes and edge types into a unified learning framework. To address this issue, \textit{meta-paths} were introduced as higher-level abstraction concepts~\cite{PathSim}. Formally, given a heterogeneous graph $\mathcal{G}=(\mathcal{V}, \mathcal{E})$, a meta-path $\Phi$ is defined as a path pattern that encodes a specific semantic relation:
\[
\Phi: Y_1 \xrightarrow{R_1} Y_2 \xrightarrow{R_2} \cdots \xrightarrow{R_\xi} Y_{\xi+1},
\]
where $Y_i \in \mathcal{Y}$ denotes a node type and $R_i \in \mathcal{R}$ denotes a relation type. Intuitively, a meta-path characterizes the composite relation between a source node of type $Y_1$ and a target node of type $Y_{\xi+1}$ by traversing a sequence of intermediate nodes and relations. Within a meta-path, the same node type may appear multiple times in the sequence, i.e., $Y_i = Y_j$ for some $i \neq j$; similarly, relation types may also repeat, i.e., $R_i = R_j$.

\begin{figure}[htbp]
\centering
{
    \begin{minipage}[b]{.98\linewidth}
    \centering
    \includegraphics[width=\linewidth]{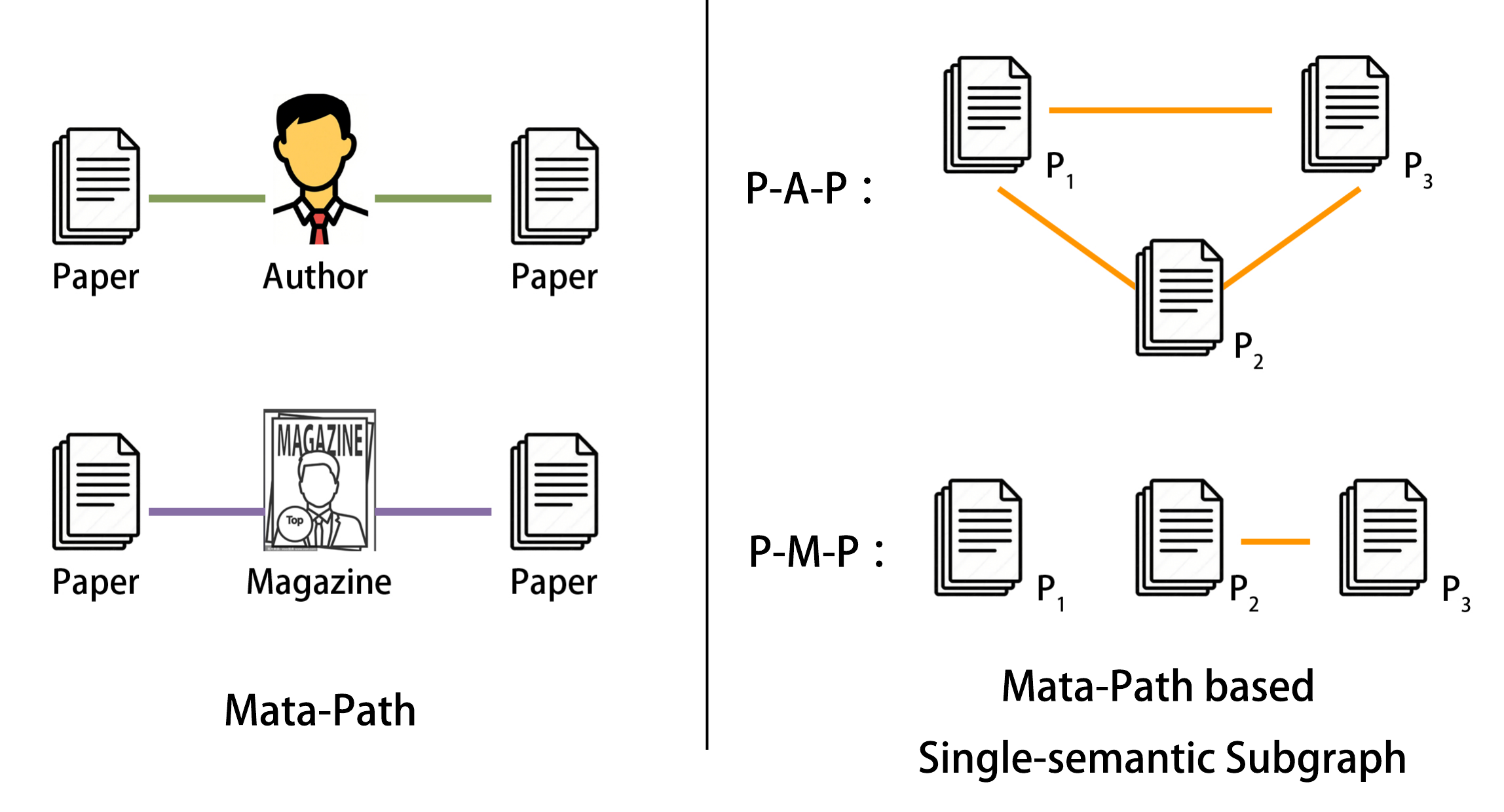}
    \end{minipage}
}
\caption{A schematic representation of the Meta-Path. Two examples of meta-paths are illustrated: Paper–Author–Paper (P–A–P) and Paper–Magazine–Paper (P–M–P). Each meta-path induces a single-semantic subgraph consisting solely of paper nodes, where edges represent semantic relations defined by the corresponding meta-path.}
\label{Schematic2}
\end{figure}

For example, taking papers as both the source and target node type, the meta-path Paper--Author--Paper (P-A-P) connects two papers that share at least one author, thereby capturing the coauthorship semantics between them. Similarly, the meta-path Paper--Magazine--Paper (P-M-P) links papers published in the same Magazine, reflecting topical or disciplinary similarity. As illustrated in Fig.~\ref{Schematic2}, these semantics-guided paths map the heterogeneous graph into single-semantic subgraphs consisting solely of source-to-target node connections,
where the edges represent the corresponding meta-path semantic relations. This transformation reduces noise from irrelevant neighbors and facilitates the learning of task-relevant semantic representations.

\subsubsection{Heterogeneous Graph Neural Networks}

Building upon the above representation methods, such as meta-paths, researchers have increasingly explored modeling paradigms that are capable of effectively capturing the complex structures and semantic information in heterogeneous graphs, thereby driving the emergence of heterogeneous graph neural networks (HGNNs). Existing HGNN models can be broadly categorized into two classes~\cite{article}
: (1) \textit{meta-path-based methods}, which explicitly construct single-semantic subgraphs via predefined meta-paths, perform feature aggregation within each subgraph, and then fuse cross-semantic information; (2) \textit{relation-based methods}, which capture implicit single-semantic weighted subgraphs among node neighbors, perform feature aggregation within each subgraph, and then achieve adaptive cross-semantic fusion.

Despite differing in their modeling approaches, both categories fundamentally follow a hierarchical semantic modeling paradigm from the semantic level to the cross-semantic level. For example, in the meta-path-based method HAN~\cite{wang2021heterogeneousgraphattentionnetwork}, semantic-level aggregation under a single meta-path is first performed. Given a target node $v_t \in \mathcal{V}$ and a meta-path $\Phi_k \in \mathcal{P}$, HAN first constructs the single-semantic subgraph $G_k = (V_k, E_k)$ derived from $\Phi_k$. For node $v_t$, its neighbors under $\Phi_k$ are denoted as $N_k(v_t)$. An attention mechanism is employed to aggregate features from these neighbors:
\[
\mathbf{h}_{v_t,\Phi_k} = \sigma \left( \sum_{u \in N_k(v_t)} a_{v_tu}^{(\Phi_k)} W^{(\Phi_k)} X_{u} \right),
\]
where $X_u$ is the feature vector of node $u$, $W^{(\Phi_k)}$ is a learnable transformation matrix specific to $\Phi_k$, $a_{v_tu}^{(\Phi_k)}$ is the attention weight that quantifies the importance of neighbor $u$ to $v_t$, and $\sigma(\cdot)$ is a nonlinear activation function.
Next, a cross-semantic-level attention mechanism is applied to combine representations across multiple meta-paths, yielding the final representation for node $v$
\[
\mathbf{h}_v = \sum_{\Phi_k \in \mathcal{P}} b_k \, \mathbf{h}_{v,\Phi_k},
\]
where $b_k$ is the attention weight reflecting the importance of the semantics represented by meta-path $\Phi_k$. The learned representation $\mathbf{h}_v$ is subsequently applied to downstream tasks such as node classification, link prediction, and complex reasoning.

Similarly, the information aggregation process in the relation-based method HetSANN~\cite{hong2019attentionbasedgraphneuralnetwork} can be decomposed into two steps: semantic-level aggregation under a single attention head and cross-semantic-level aggregation across multiple attention heads. The former captures an implicit single-semantic subgraph through a single attention head and performs information aggregation. The latter then concatenates the diverse single-semantic information from multiple attention heads to achieve cross-semantic-level aggregation, which can subsequently be fine-tuned for specific tasks. Therefore, whether meta-path-based or relation-based, the core idea of both categories follows the same paradigm: semantic-level aggregation first, followed by cross-semantic-level fusion.

\subsection{Problem Statement}
\label{sec:problem-statement}

Given this hierarchical semantic structure, we formalize the interpretability task for HGNNs as providing explanations at both the semantic level and the cross-semantic level.

Formally, given a heterogeneous graph $\mathcal{G} = (\mathcal{V}, \mathcal{E})$ with node type mapping $\phi: \mathcal{V} \rightarrow \mathcal{Y}$, edge type mapping $\psi: \mathcal{E} \rightarrow \mathcal{R}$, a set of meta-paths $\mathcal{P} = \{\Phi_1, \Phi_2, \dots, \Phi_m\}$, a node feature matrix $X \in \mathbb{R}^{|\mathcal{V}| \times n}$, a model $M$, and a target node $v_t \in \mathcal{V}$, we aim to learn an explanation function $F$:
\[
(\mathcal{G}, \mathcal{P}, X, M, v_t) \mapsto (\mathbf{S}, \mathbf{c})
\]
where $\mathbf{S} \in \mathbb{R}^{m \times n}$ is a semantic explanation matrix whose $k$-th row $\bm{s}_k$ denotes the feature importance under meta-path $\Phi_k$, and $\mathbf{c} \in \mathbb{R}^n$ is a cross-semantic explanation vector that integrates all the semantics to yield the overall feature importance for the prediction $M(\mathcal{G}, X, v_t)$.

Although our task formulation adopts the concept of meta-paths, for relation-based methods, the single-semantic subgraphs learned by the model can be expressed in a meta-path-based form to maintain consistency within this framework.

\begin{figure*}[!htbp]
    \centering
    \includegraphics[width=0.98\textwidth]{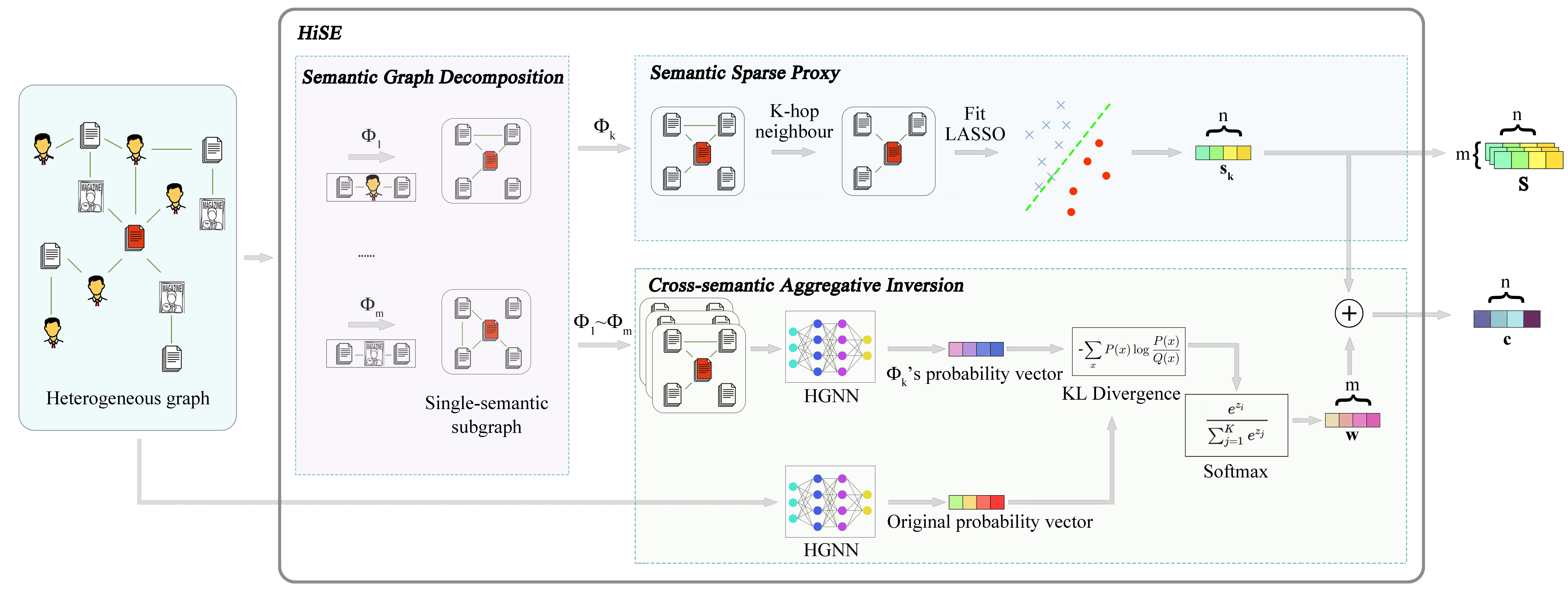}
    \caption{Overview of the HiSE framework. The heterogeneous graph is decomposed into single-semantic subgraphs. Within each subgraph, the SSP component fits a weighted sparse surrogate model via LASSO to obtain per-semantic feature explanations $\bm{s}_k$, which collectively form the semantic-level explanation $\mathbf{S}$. The CAI component then evaluates the contribution of each semantic view by comparing the prediction distributions of these subgraphs against that of the full graph via KL divergence, yielding semantic weights $\boldsymbol{w}$. With these weights, CAI aggregates $\bm{s}_k$ into the cross-semantic-level explanation $\mathbf{c}$.}

    \label{Schematic4}
\end{figure*}

\section{Methodology}

To provide more effective hierarchical semantic explanations for HGNNs, we propose the \textbf{HiSE framework}, comprising two core components: \textbf{semantic sparse proxy (SSP)} and \textbf{cross-semantic aggregative inversion (CAI)}. First, the heterogeneous graph is decomposed into multiple single-semantic subgraphs on the basis of predefined meta-paths, serving as the shared input for both components. On this basis, in the SSP component, sparse surrogate models are constructed at each semantic view to reveal feature contributions, thereby obtaining semantic-level feature explanations for each semantic; in the CAI component, the contribution of each semantic context to model decisions is inverted, yielding a unified cross-semantic feature explanation through weighted integration. The overall workflow of the HiSE framework is illustrated in Fig.~\ref{Schematic4}, and its overall procedure is presented in the pseudocode in the Appendix.

\subsection{Semantic Graph Decomposition}

On the basis of a predefined set of meta-paths $\mathcal{P}=\{\Phi_1, \Phi_2, \ldots, \Phi_m\}$, we perform semantic decomposition on the original heterogeneous graph $\mathcal{G} = (\mathcal{V}, \mathcal{E})$ to obtain a set of single-semantic subgraphs $\mathbb{G} = \{G_1, G_2, \ldots, G_m\}$ corresponding to the meta-paths. Given a meta-path $\Phi_k$: \(
v_1 \xrightarrow{r_1} v_2 \xrightarrow{r_2} \cdots \xrightarrow{r_\xi} v_{\xi+1}
\), where $v_i \in \mathcal{V}$ and $r_j \in \mathcal{E}$, the adjacency matrix of the single-semantic subgraph corresponding to $\Phi_k$ is computed as follows:
\begin{equation}
\mathbf{A}_{\Phi_k} = \mathbf{A}_{r_1} \times \mathbf{A}_{r_2} \times \cdots \times \mathbf{A}_{r_\xi},
\label{eq:adj_product}
\end{equation}
where $\mathbf{A}_{r_i}$ denotes the adjacency matrix under relation type $r_i$, with $\mathbf{A}_{r_i}(u, v) = 1$ if and only if there exists an edge $(u, v)$ of type $r_i$ in $\mathcal{G}$, and $\times$ denotes the matrix outer product. The single-semantic subgraph corresponding to $\Phi_k$ is then constructed as follows:
\begin{equation}
{G}_k = ({V}_k, {E}_k),
\label{eq:subgraph}
\end{equation}
where ${E}_k = \{(u, v) \mid \mathbf{A}_{\Phi_k}(u, v) > 0\}$, and ${V}_k$ denotes the set of all nodes reachable through ${E}_k$.

\subsection{Semantic Sparse Proxy}
In this module, we aim to generate independent feature explanations $\bm{s}_k \in \mathbb{R}^n$ for the target node $v_t$ on each single-semantic subgraph $G_k$, thereby forming the semantic explanation matrix $\mathbf{S} \in \mathbb{R}^{m \times n}$. 
The core idea is to construct a sparse surrogate model at each semantic view to fit the prediction behavior of $M$, thereby yielding explanations specific to that semantic view.

\subsubsection{$K$-Hop Weighted Neighborhood Sampling}
For a given meta-path $\Phi_k$ and its corresponding single-semantic subgraph ${G}_k$, we construct a local weighted neighborhood dataset $\mathcal{D}_k$ for explanation by extracting the $K$-hop neighborhood structure of the target node $v_t$ in ${G}_k$ and assigning semantically relevant weights to nodes within the neighborhood. Given a hyperparameter $K$, the corresponding neighborhood node set is defined as follows:
\begin{equation}
{N}_k = \{ v_i \mid v_i \in \mathcal{V}_k,\
\text{hop-distance}_{G_k}(v_t, v_i) \leq K \}.
\label{eq:neighborhood}
\end{equation}

Under the semantic constraint of meta-path $\Phi_k$, for any neighborhood node $v_i \in N_k$, we define its unnormalized weight $\tilde{\mu}_i^{(k)}$ as the sum of edge weight products across all paths from $v_t$ to $v_i$ that conform to $\Phi_k$:
\begin{equation}
\tilde{\mu}_i^{(k)} =
\sum_{p \in \mathcal{P}_{\Phi_k}(v_t, v_i)}
\prod_{(u,v)\in p} \eta_{uv},
\label{eq:raw_weight}
\end{equation}
where $\mathcal{P}_{\Phi_k}(v_t, v_i)$ denotes the set of all paths from $v_t$ to $v_i$ in the single-semantic subgraph ${G}_k$ that conform to meta-path $\Phi_k$ and $\eta_{uv}$ denotes the weight assigned to edge $(u,v)$ by the model to be explained.

The weights of all the nodes in the neighborhood are subsequently normalized and further scaled to maintain the numerical stability of the loss function:
\begin{equation}
\mu_i^{(k)} =
\frac{\tilde{\mu}_i^{(k)}}{\sum_{v_j \in N_k} \tilde{\mu}_j^{(k)}}
\cdot |N_k| .
\label{eq:norm_weight}
\end{equation}
For each node $v_i \in N_k$, we feed its feature vector $X_i \in \mathbb{R}^n$ along with the original heterogeneous graph $\mathcal{G}$ as input to obtain the prediction probability distribution of the model $M$ to be explained:
\begin{equation}
\bm{\hat{y}}^{\mathcal{G}}_i \gets M(\mathcal{G}, X_i) \label{eq:prediction}
\end{equation}
where $\bm{\hat{y}}^{\mathcal{G}}_i \in \mathbb{R}^L$ is the predicted probability vector over $L$ classes and $\hat{y}^{\mathcal{G}}_{i,l}$ denotes the predicted probability for the $l$-th class.

Finally, the weighted local neighborhood dataset is constructed as follows:
\begin{equation}
\mathcal{D}_k =
\{ (X_i, \bm{\hat{y}}^{\mathcal{G}}_i, \mu_i^{(k)}) \mid v_i \in N_k \}.
\label{eq:dataset}
\end{equation}

\subsubsection{Weighted Sparse Proxy Fitting}
On the basis of the weighted local neighborhood dataset $\mathcal{D}_k$, we learn a sparse linear surrogate model for each semantic $\Phi_k$:
\begin{equation}
g_k(X)=\bm{s}_k^\top X ,
\end{equation}
thereby obtaining the semantic-level feature explanation $\bm{s}_k\in\mathbb{R}^n$ for that semantic. To fully account for the differences in importance among sampled nodes under semantic $\Phi_k$, we introduce sample weights $\mu_i^{(k)}$ and employ the weighted Least Absolute Shrinkage and Selection Operator via Least Angle Regression (LassoLARS) method~\cite{Efron_2004,Tibshirani1996RegressionSA} to optimize the sparse linear surrogate model $g_k(X)$ to fit $\mathcal{D}_k$.

For all $(X_i,\bm{\hat{y}}^{\mathcal{G}}_i,\mu_i^{(k)}) \in \mathcal{D}_k$, LassoLARS minimizes the following weighted LASSO objective:
\begin{equation}
\mathcal{L}(\bm{s}_{k,l})
=
\sum_{v_i \in N_k}
\mu_i^{(k)}
\left(
\hat{y}^{\mathcal{G}}_{i,l}
-
\bm{s}_{k,l}^\top X_i
\right)^2
+
\lambda \|\bm{s}_{k,l}\|_1 ,
\label{eq:lasso}
\end{equation}
where $\bm{s}_{k,l}\in\mathbb{R}^n$ denotes the semantic-level feature explanation for class $l$ under semantic $\Phi_k$ and $\mu_i^{(k)}$ represents the importance weight of node $v_i$ under the current semantic.

This optimization is solved via the LassoLARS algorithm (detailed in Appendix), which is performed independently for each class $l$.

To reflect the overall prediction behavior, we average the explanation results across all classes:
\begin{equation}
\bm{s}_k
=
\frac{1}{L}
\sum_{l=1}^L
\bm{s}^{(\text{final})}_{k,l}.
\label{eq:semantic_exp}
\end{equation}

Repeating this process for all meta-paths $\Phi_k \in \mathcal{P}$ yields a set of semantic-level feature explanations
\[
\mathbf{S} = \{\bm{s}_1,\bm{s}_2,\ldots,\bm{s}_m\},
\]
that characterize the key features influencing model predictions from different semantic views. The complete procedure is summarized in the pseudocode in the Appendix.

\subsection{Cross-Semantic Aggregative Inversion}

In this module, we evaluate the relative contribution $w_k$ of each semantic meta-path $\Phi_k$ to the final prediction in a model-agnostic manner and synthesize a cross-semantic explanation $\mathbf{c} \in \mathbb{R}^n$ by combining the semantic-level explanations $\bm{s}_k$ from SSP. The core idea is to measure the consistency between the model's predictions on the single-semantic subgraph $G_k$ and those on the full graph $\mathcal{G}$.

Specifically, we first obtain the global prediction distribution $\bm{\hat{y}}^{\mathcal{G}} = [\hat{y}^{\mathcal{G}}_1, \hat{y}^{\mathcal{G}}_2, \ldots, \hat{y}^{\mathcal{G}}_{|\mathcal{N}|}]$ by feeding $X$ and $\mathcal{G}$ into the model. For each meta-path $\Phi_k$, we similarly obtain the prediction distribution $\bm{\hat{y}}^{G_k}$ using the subgraph $G_k$. The negative KL divergence between the two distributions serves as the similarity score $\gamma_k$:
\begin{equation}
\gamma_k = - \sum_{v_i \in \mathcal{N}} \bm{\hat{y}}^{\mathcal{G}}_i
\log \frac{\bm{\hat{y}}^{\mathcal{G}}_i}{\bm{\hat{y}}^{G_k}_i}
\label{eq:kl_div}
\end{equation}
where a higher $\gamma_k$ indicates greater consistency, meaning that $\Phi_k$ contributes more critical semantic information to the model's decisions.

The similarity scores $\bm{\gamma} = [\gamma_1, \gamma_2, \ldots, \gamma_m]$ are then normalized into semantic weights $\boldsymbol{w} = [{w_1}, {w_2}, \ldots, {w_m}]$ via softmax:
\begin{equation}
{w_k} = \frac{\exp(\gamma_k)}{\sum_{k=1}^{m} \exp(\gamma_k)}
\label{eq:softmax_beta}
\end{equation}

Finally, the cross-semantic explanation for the target node $v_t$ is obtained as follows:
\begin{equation}
\mathbf{c} = \sum_{k=1}^{m} {w_k} \cdot \bm{s}_k
\label{eq:final_weight}
\end{equation}
where $\mathbf{c} \in \mathbb{R}^n$ aggregates feature contributions across all meta-paths, providing a comprehensive answer to which features are most influential and through which semantic relationships they operate. The complete procedure is summarized in the pseudocode in the Appendix.

\section{Experiments}

To comprehensively evaluate the performance of HiSE, we design a series of experiments from two perspectives: effectiveness and efficiency. The effectiveness evaluation covers four key aspects: explanation accuracy, robustness, usability, and cross-semantic explanation capability; the efficiency evaluation focuses on computational overhead.

\subsection{Heterogeneous Graph Neural Networks to be Explained}

We adopt two representative meta-path-based HGNNs and two representative relation-based HGNNs as target models for explanation:

\subsubsection{\textbf{Meta-Path-Based Methods.}}
The meta-path-based methods rely on manually designed meta-paths to capture high-order semantic structures and perform message passing on the induced single-semantic subgraphs.

\begin{itemize}

    \item \textbf{HAN~\cite{wang2021heterogeneousgraphattentionnetwork}} introduces a hierarchical attention mechanism with node-level (per meta-path) and semantic-level (cross meta-path) attention, which correspond to the semantic level and cross-semantic level, respectively, in this work.

    \item \textbf{MAGNN~\cite{Fu_2020}} extends meta-path-based message passing by incorporating intermediate nodes and aggregating multiple meta-path instances to capture richer contextual semantics, combining both intra- and intermeta-path aggregation.
\end{itemize}

\subsubsection{\textbf{Relation-Based Methods.}}
The relation-based methods directly use relation types as the basic modeling unit and learn semantic structures via relation-aware message passing, avoiding the need for manual meta-path design.

\begin{itemize}
    \item \textbf{HetSANN~\cite{hong2019attentionbasedgraphneuralnetwork}} employs relation-specific attention to aggregate neighbor information from different relation types within the same layer, capturing diverse connectivity patterns without predefined meta-paths.

    \item \textbf{HGT~\cite{10.1145/3366423.3380027}} introduces type-aware projection matrices and attention parameters for different node and relation types, enabling relation-level structured attention within a multihead framework to model high-order interactions in heterogeneous graphs.
\end{itemize}

\subsection{Datasets}

We use two widely adopted academic network datasets in our experiments, with detailed statistics provided in the Appendix.

\begin{itemize}
    \item \textbf{ACM~\cite{wang2021heterogeneousgraphattentionnetwork}.}
    Sourced from papers published at KDD, SIGMOD, SIGCOMM, MobiCOMM, and VLDB, the data in the ACM dataset are categorized into three classes (\textit{database}, \textit{wireless communication}, \textit{data mining}). Paper features are represented as bag-of-words vectors of keywords, and labels are assigned on the basis of the publishing conference.

    \item \textbf{MAG.}
    The MAG dataset was constructed by sampling from \textit{ogbn-mag}~\cite{ogbn_mag} to form a heterogeneous academic network subgraph. We randomly select 5,000 paper nodes belonging to three classes and retain all connected author and field-of-study nodes and their relations. Papers include 128-dimensional word2vec features.
\end{itemize}

\subsection{Baseline Methods}

We compare HiSE against baselines from three categories: basic methods, such as performance lower bounds; homogeneous GNN explanation methods, to examine their applicability in heterogeneous scenarios; and heterogeneous GNN explanation methods, which are the key comparative objects.

\textbf{(1) Basic Methods.}
\begin{itemize}
    \item \textbf{Random}~\cite{ribeiro2016whyitrustyou}: Randomly selects features under a given budget, providing a random lower bound.
    \item \textbf{Greedy}~\cite{10.25300/MISQ/2014/38.1.04}: Sequentially selects features in descending order of absolute values.
\end{itemize}

\textbf{(2) Homogeneous GNN Explanation Methods.}
For fair comparison, we construct homogeneous subgraphs on the basis of predefined meta-paths and randomly select one as the input for each method.
\begin{itemize}
    \item \textbf{GraphLIME}~\cite{huang2020graphlimelocalinterpretablemodel}: Extends LIME~\cite{ribeiro2016whyitrustyou} to graphs by learning nonlinear surrogate models in local neighborhoods.
    \item \textbf{GNNExplainer}~\cite{ying2019gnnexplainergeneratingexplanationsgraph}: Identifies critical subgraphs and features by optimizing soft masks to maximize mutual information with model predictions.
    \item \textbf{ZORRO}~\cite{funke2021hard}: Progressively selects features that maximize fidelity, measured by prediction consistency after the selected features are fixed and the remaining features are replaced with random noise.
\end{itemize}

\textbf{(3) Heterogeneous GNN Explanation Methods.}
\begin{itemize}
    \item \textbf{HENCE-X}~\cite{HENCEX}: Employs Markov blanket search to identify the most relevant feature subset and then constructs explanations via statistical significance testing.
    \item \textbf{HGExplainer}~\cite{HGExplainer}: Extends GNNExplainer to heterogeneous graphs by learning structural and feature masks on meta-path-induced subgraphs and aggregating them.
\end{itemize}

\subsection{Experimental Setup}
\label{subsec:exp_settings}

\subsubsection{Explainer Settings}
\begin{itemize}
    \item \textbf{HiSE}: For meta-path-based models (HAN and MAGNN), each predefined meta-path is treated as an independent semantic unit. For relation-based models (HetSANN and HGT), each attention head serves as an independent semantic unit.
 
    \item \textbf{HENCE-X}~\cite{HENCEX}: We limit the upper bound of neighbor sampling per type to 50 and the maximum number of perturbed samples to 2,000 to ensure computational feasibility.

    \item \textbf{ZORRO}~\cite{funke2021hard}: We limit the number of random samplings to 20 to reduce the computational overhead.

    \item \textbf{Other baselines}~\cite{ribeiro2016whyitrustyou, 10.25300/MISQ/2014/38.1.04, huang2020graphlimelocalinterpretablemodel, ying2019gnnexplainergeneratingexplanationsgraph, HGExplainer}: We adopt the original implementations.
\end{itemize}

\subsubsection{Noise-Augmented Settings}

We conduct experiments on the ACM and MAG datasets. To evaluate the fidelity, robustness, and usability of each explainer under irrelevant feature interference, we append 30\% random noise features to the original node features. For the ACM dataset (binary features), we generate i.i.d. random binary noise; for the MAG dataset (continuous features), we generate uniform noise over $[-1, 1]$. The noise features are concatenated with the original features and are independent of class labels. All the HGNN classifiers are trained on the complete feature space with noise appended.

\subsection{\textbf{Experiment 1 Fidelity}: How Trustworthy is the Explainer?}

\begin{table*}[htbp]
\centering
\caption{Fidelity comparison of eight explainers under different HGNN classifiers, 
datasets, and explanation budgets $K$. Higher values indicate better fidelity ($\uparrow$). 
The best and second-best results in each setting are highlighted in \textbf{bold} and \underline{underlined}, respectively.}
\label{tab:trustworthiness_main}
\setlength{\tabcolsep}{4pt}
\renewcommand{\arraystretch}{1.1}
\begin{tabular}{c c c cccccccc}
\toprule
\multirow{3}{*}{\textbf{Classifier}}
& \multirow{3}{*}{\textbf{Dataset}}
& \multirow{3}{*}{\textbf{$K$}}
& \multicolumn{8}{c}{\textbf{Explainer  (Fidelity $\uparrow$)}} \\

\cmidrule(lr){4-11}
& & 
& \multicolumn{2}{c}{\textbf{Basic}}
& \multicolumn{3}{c}{\textbf{Homogeneous}}
& \multicolumn{2}{c}{\textbf{Heterogeneous}}
& \multicolumn{1}{c}{\textbf{Ours}} \\

\cmidrule(lr){4-5} \cmidrule(lr){6-8} \cmidrule(lr){9-10} \cmidrule(lr){11-11}
& & 
& Random & Greedy 
& GraphLIME & ZORRO & GNNExplainer 
& HGExplainer & HENCE-X 
& HiSE \\

\midrule

\multirow{8}{*}{\raisebox{9\height}{HAN}}
& \multirow{4}{*}{\raisebox{4\height}{ACM}}
& 10 &            0.70& 0.71& \textbf{0.85}& 0.81& 0.68& 0.71& 0.78& \underline{0.82}\\
& & 20 & 0.74& 0.74& \underline{0.87}& 0.84& 0.72& 0.67& 0.81& \textbf{0.89}\\
& \multirow{4}{*}{\raisebox{4.5\height}{MAG}}
& 10 & 0.74 & 0.52 & 0.81 & 0.81 & 0.82& \underline{0.85}& 0.57 & \textbf{0.86}\\
& & 20 & 0.85 & 0.52 & 0.83 & 0.84 & \underline{0.89}& 0.88 & 0.74 & \textbf{0.91}\\
\midrule

\multirow{8}{*}{\raisebox{9\height}{MAGNN}}
& \multirow{4}{*}{\raisebox{4\height}{ACM}}
& 10 & 0.72 & 0.74 & 0.77& \underline{0.81}& 0.73 & 0.73 & 0.73 & \textbf{0.83}\\
& & 20 & 0.74 & 0.78 & 0.80 & \underline{0.82} & 0.76 & 0.69 & 0.77 & \textbf{0.86} \\
& \multirow{4}{*}{\raisebox{4.5\height}{MAG}}
& 10 & 0.72& 0.76& 0.75& \underline{0.78}& 0.73& 0.74& 0.74& \textbf{0.81}\\
& & 20 & 0.76& \underline{0.83}& 0.82& \textbf{0.84}& 0.74& 0.78& 0.82& 0.82\\
\midrule

\multirow{8}{*}{\raisebox{9\height}{HetSANN}}
& \multirow{4}{*}{\raisebox{4\height}{ACM}}
& 10 & 0.52 & 0.52 & \underline{0.66} & 0.54 & 0.52 & 0.51 & 0.50 & \textbf{0.78} \\
& & 20 & 0.52 & 0.51 & \underline{0.65} & 0.55 & 0.51 & 0.53 & 0.54 & \textbf{0.72} \\
& \multirow{4}{*}{\raisebox{4.5\height}{MAG}}
& 10 & 0.80& 0.54& 0.84& 0.79 & 0.80& \underline{0.90}& 0.76& \textbf{0.94}\\
& & 20 & 0.89& 0.44& 0.91& 0.87& 0.92& \textbf{0.97}& 0.87& \underline{0.95}\\
\midrule

\multirow{8}{*}{\raisebox{9\height}{HGT}}
& \multirow{4}{*}{\raisebox{4\height}{ACM}}
& 10 & 0.72 & 0.71 & 0.76 & 0.71 & 0.73 & 0.69 & \underline{0.78} & \textbf{0.86} \\
& & 20 & 0.74 & 0.73 & \underline{0.79} & 0.74 & 0.73 & 0.70 & \underline{0.79} & \textbf{0.89} \\
& \multirow{4}{*}{\raisebox{4.5\height}{MAG}}
& 10 & 0.47 & 0.34 & \underline{0.62} & 0.59 & 0.49 & 0.51 & 0.36 & \textbf{0.64} \\
& & 20 & 0.58 & 0.38 & \underline{0.66} & 0.61 & 0.64 & 0.59 & 0.50 & \textbf{0.68} \\
\bottomrule
\end{tabular}
\end{table*}

A faithful explanation should accurately identify the features on which the model truly relies. To evaluate this fidelity, we first train an HGNN classifier with accuracy $\geq 85\%$ on the noise-augmented feature space. For a given node $v_i$, the explainer selects the top-$K$ most important features $\mathcal{F}_i^{(K)}$. We then retrain $\Gamma=10$ classifiers using only these features under the same data split and measure prediction consistency with the original model:
\begin{equation}
\mathrm{Fid}_i =
\frac{1}{\Gamma} \sum_{\tau=1}^{\Gamma}
\mathbb{I}\!\left( f^{(\tau)}(v_i \mid \mathcal{F}_i^{(K)}) = f(v_i) \right),
\end{equation}
where $f(\cdot)$ and $f^{(\tau)}(\cdot)$ denote the original and $\tau$-th retrained models, respectively, and $\mathbb{I}(\cdot)$ is the indicator function. The overall fidelity is averaged over 200 randomly selected test nodes:
\begin{equation}
\mathrm{Fidelity} = \frac{1}{|\mathcal{V}|} \sum_{v_i \in \mathcal{V}} \mathrm{Fid}_i.
\end{equation}

The results are presented in Table~\ref{tab:trustworthiness_main}. HiSE achieves the highest and most stable fidelity across nearly all the settings. HENCE-X exhibits limited and unstable performance. HGExplainer, ZORRO, and GraphLIME achieve reasonable results in some scenarios but remain inconsistent overall.

The advantage of HiSE stems from its semantic-level surrogate modeling and cross-semantic aggregation, which accurately reflect the semantic propagation mechanism of HGNNs and stably identify influential features even under noise. In contrast, GraphLIME, GNNExplainer, and ZORRO were designed for homogeneous graphs and cannot explicitly capture multisemantic propagation structures in HGNNs. Although HGExplainer and HENCE-X target heterogeneous graphs, they cannot faithfully reflect the hierarchical semantic mechanism underlying HGNNs and fail to accurately identify the features that truly drive model predictions.

\subsection{\textbf{Experiment 2 Robustness}: How Resistant is the Explainer to Noise?}

\begin{figure*}[htbp]
\centering
\subfloat[HAN on ACM]{
    \includegraphics[width=0.22\linewidth]{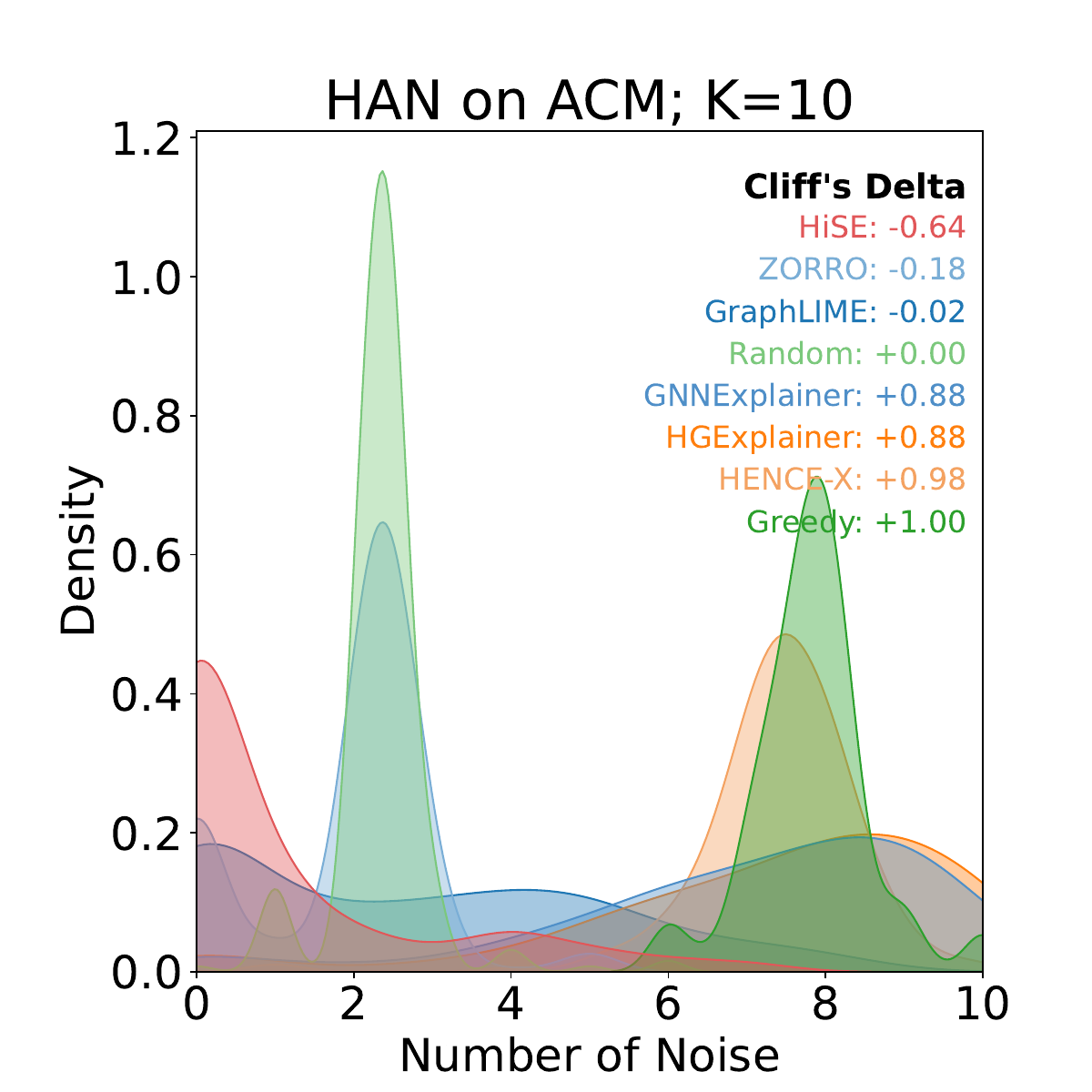}
}
\hfill
\subfloat[MAGNN on ACM]{
    \includegraphics[width=0.22\linewidth]{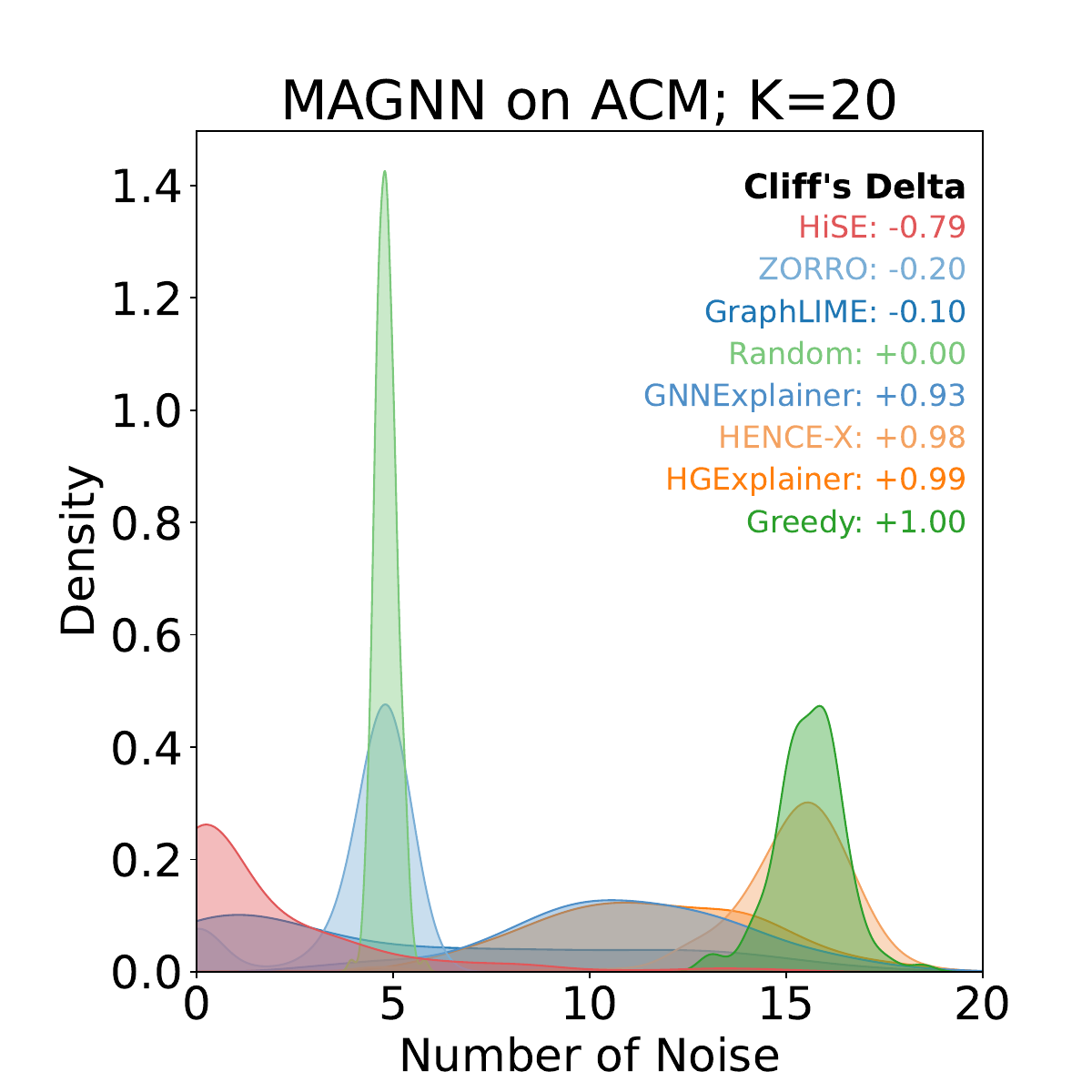}
}
\hfill
\subfloat[HetSANN on MAG]{
    \includegraphics[width=0.22\linewidth]{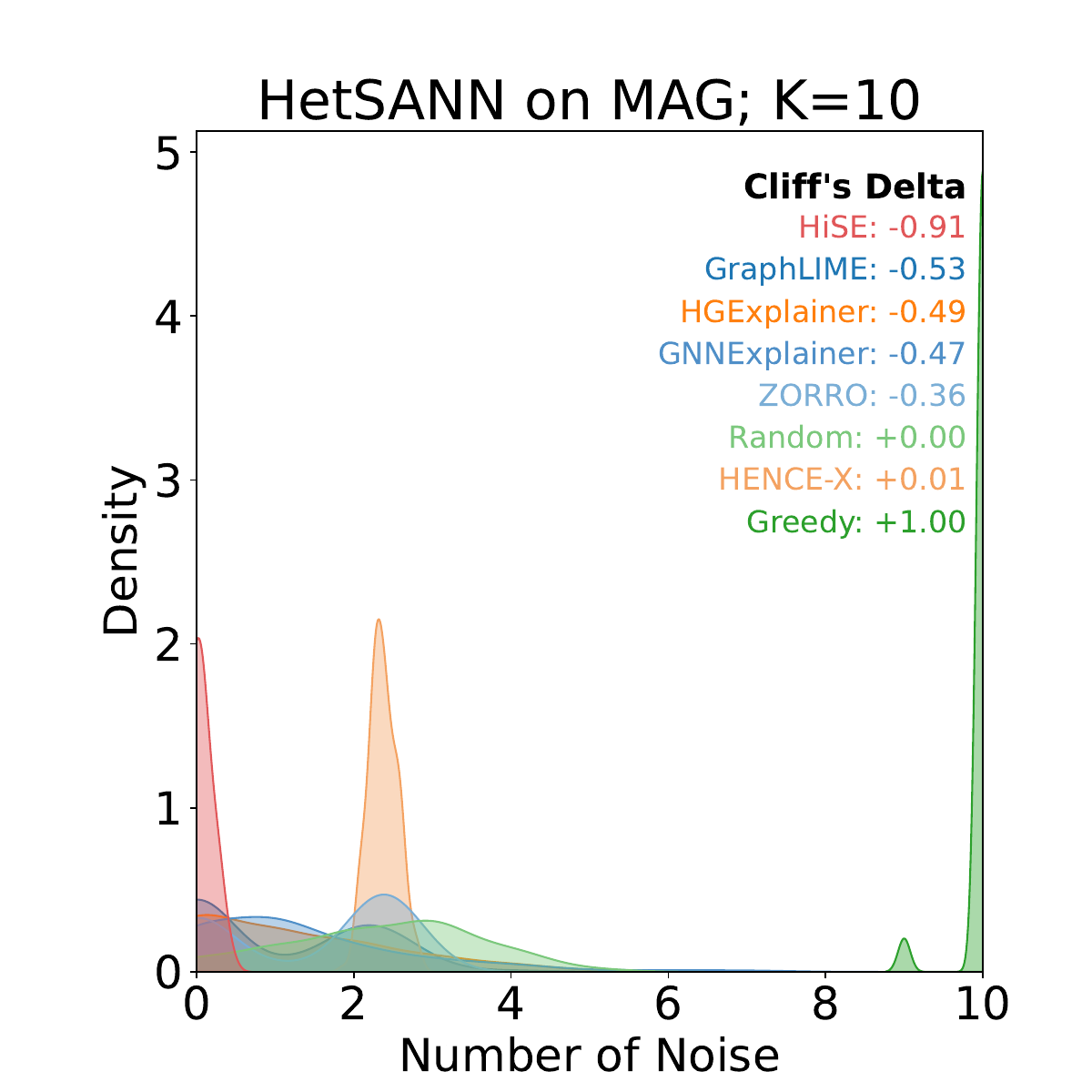}
}
\hfill
\subfloat[HGT on MAG]{
    \includegraphics[width=0.22\linewidth]{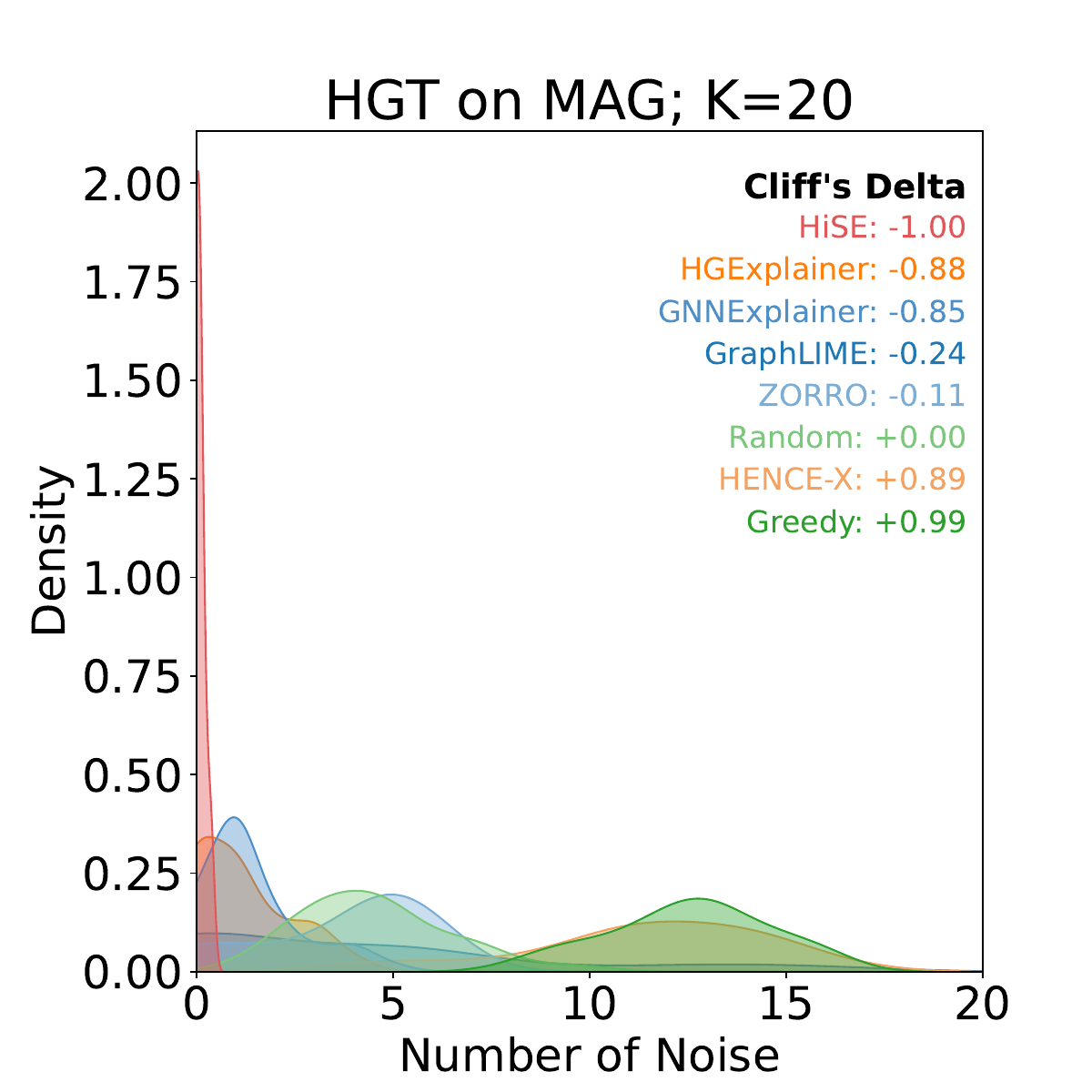}
}

\vspace{0.5em}
\includegraphics[width=1\linewidth]{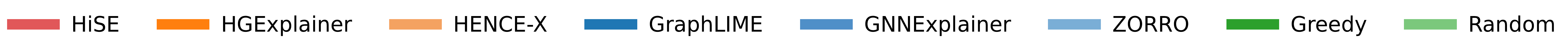}

\caption{Frequency distribution of noise features selected by different explainers, visualized using KDE curves. 
Each subplot corresponds to a different HGNN model--dataset--$K$ setting. 
A lower concentration indicates stronger explainer robustness to noise.
Complete results across all settings are deferred to Appendix.}
\label{num_noise_features}
\end{figure*}

Following the noise-augmented setting in Section~\ref{subsec:exp_settings}, we evaluate the robustness of each explainer using HGNN classifiers with test accuracy exceeding 85\%. For 200 randomly selected test nodes, each explainer generates the top-$K$ ($K=10,20$) feature explanations, and we count the number of noise features $N$ among the selected features, where a smaller $N$ indicates stronger robustness. We repeat the process 10 times and plot the distribution of the $N$ values. Additionally, we compute Cliff's delta effect size~\cite{Cliffs_delta} for each method relative to Random as a quantitative measure of noise suppression capability. The results are shown in Fig.~\ref{num_noise_features}.

HiSE demonstrates the strongest noise suppression capability, with its $N$ distribution significantly lower than that of Random and achieving the best Cliff's delta across the vast majority of settings. Greedy performs worst across nearly all the scenarios. HENCE-X, HGExplainer, and GNNExplainer exhibit poor robustness, with $N$ distributions notably higher than those of Random in many cases, indicating that they frequently misidentify noise features as important. GraphLIME and ZORRO show moderate robustness, slightly outperforming Random.

This advantage stems from HiSE's surrogate model-based paradigm, which learns an interpretable linear model to approximate the original model's decision boundary without perturbing input features, thereby preserving the original feature distribution and effectively suppressing noise. In contrast, perturbation-based methods (HENCE-X, HGExplainer, and GNNExplainer) evaluate importance by masking or removing input features, which tends to disrupt the original distribution and amplify the incidental influence of noise in high-dimensional feature spaces, leading to larger $N$ values.

\subsection{\textbf{Experiment 3 Usability}: Can the Explainer Help Identify Better Models?}

\label{subsec:choose_better_model}
\begin{figure*}[htbp]
\centering
\subfloat[HAN on ACM]
{
    \begin{minipage}[b]{0.22\linewidth}
    \centering
    \includegraphics[width=\linewidth]{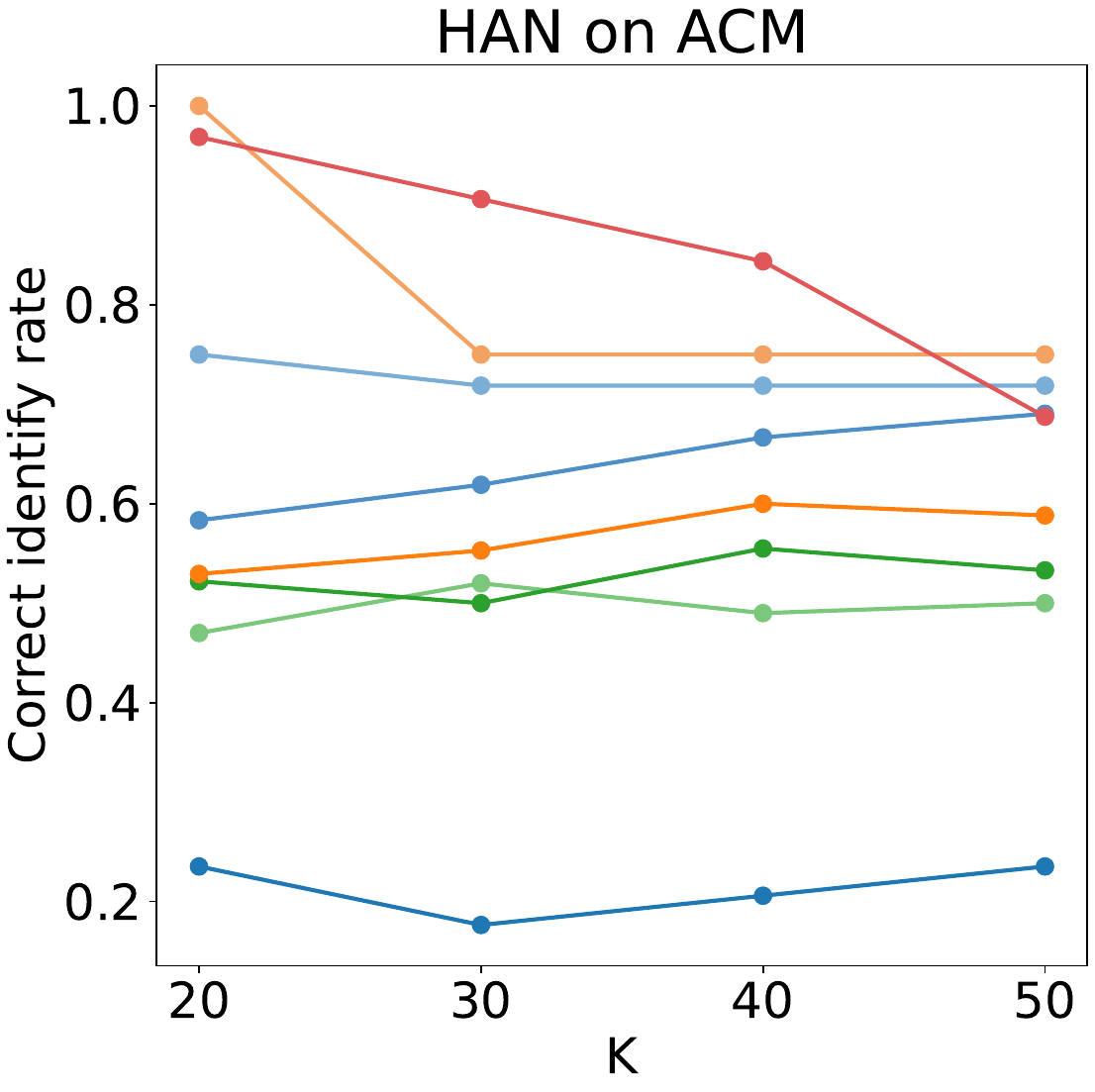}
    \end{minipage}
}
\subfloat[MAGNN on ACM]
{
    \begin{minipage}[b]{0.22\linewidth}
    \centering
    \includegraphics[width=\linewidth]{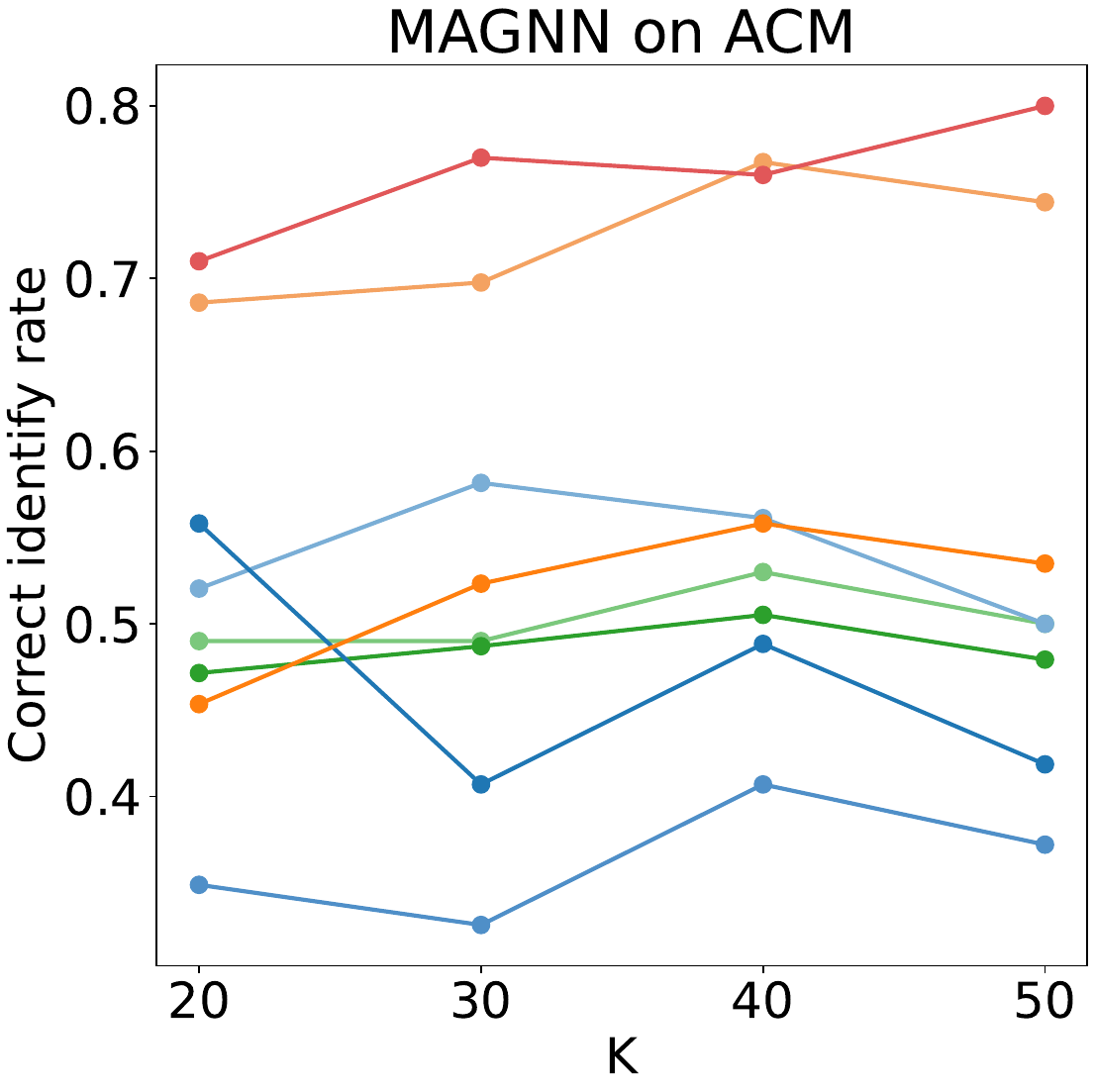}
    \end{minipage}
}
\subfloat[HetSANN on MAG]
{
    \begin{minipage}[b]{0.22\linewidth}
    \centering
    \includegraphics[width=\linewidth]{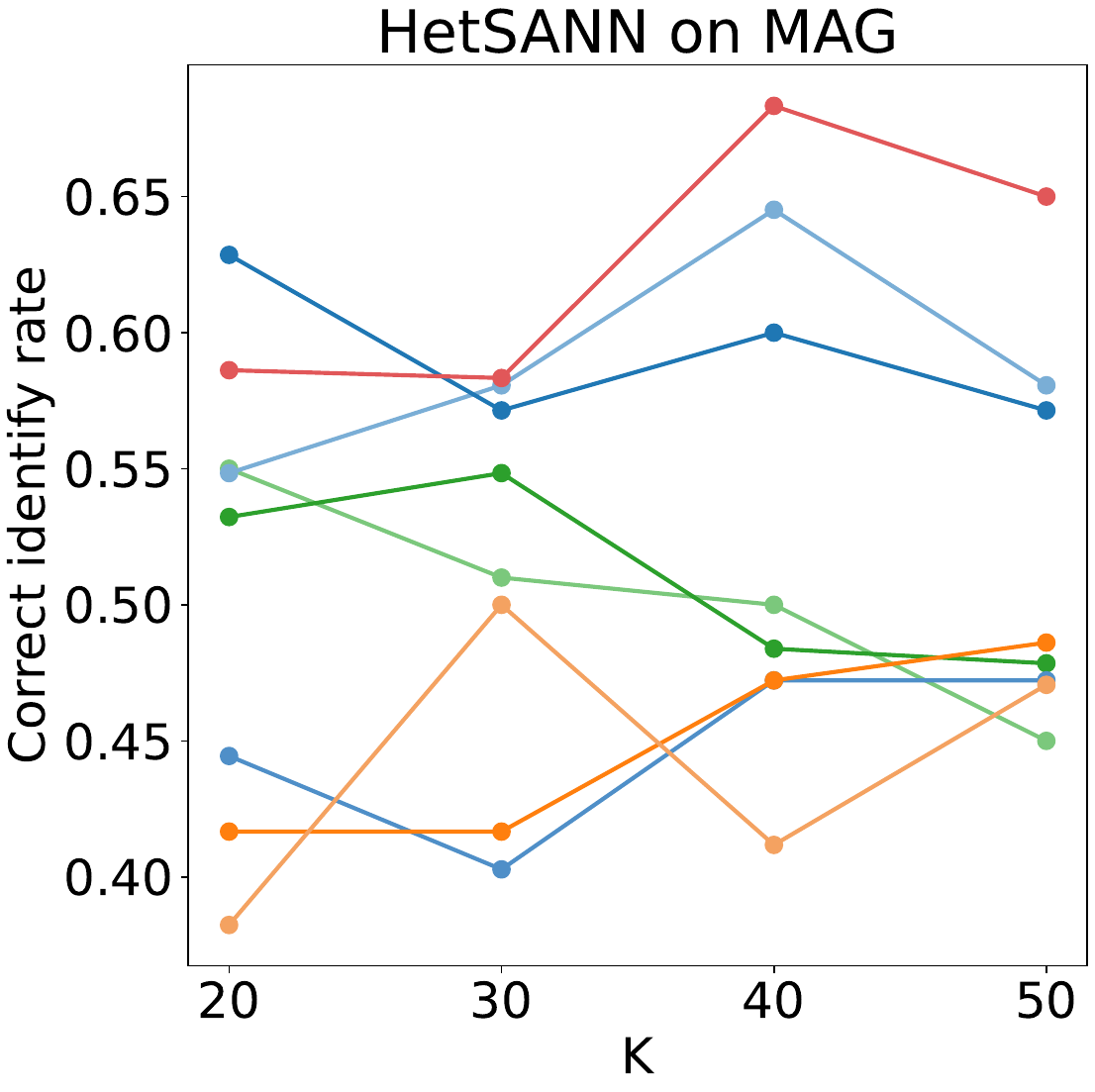}
    \end{minipage}
}
\subfloat[HGT on MAG]
{
    \begin{minipage}[b]{0.22\linewidth}
    \centering
    \includegraphics[width=\linewidth]{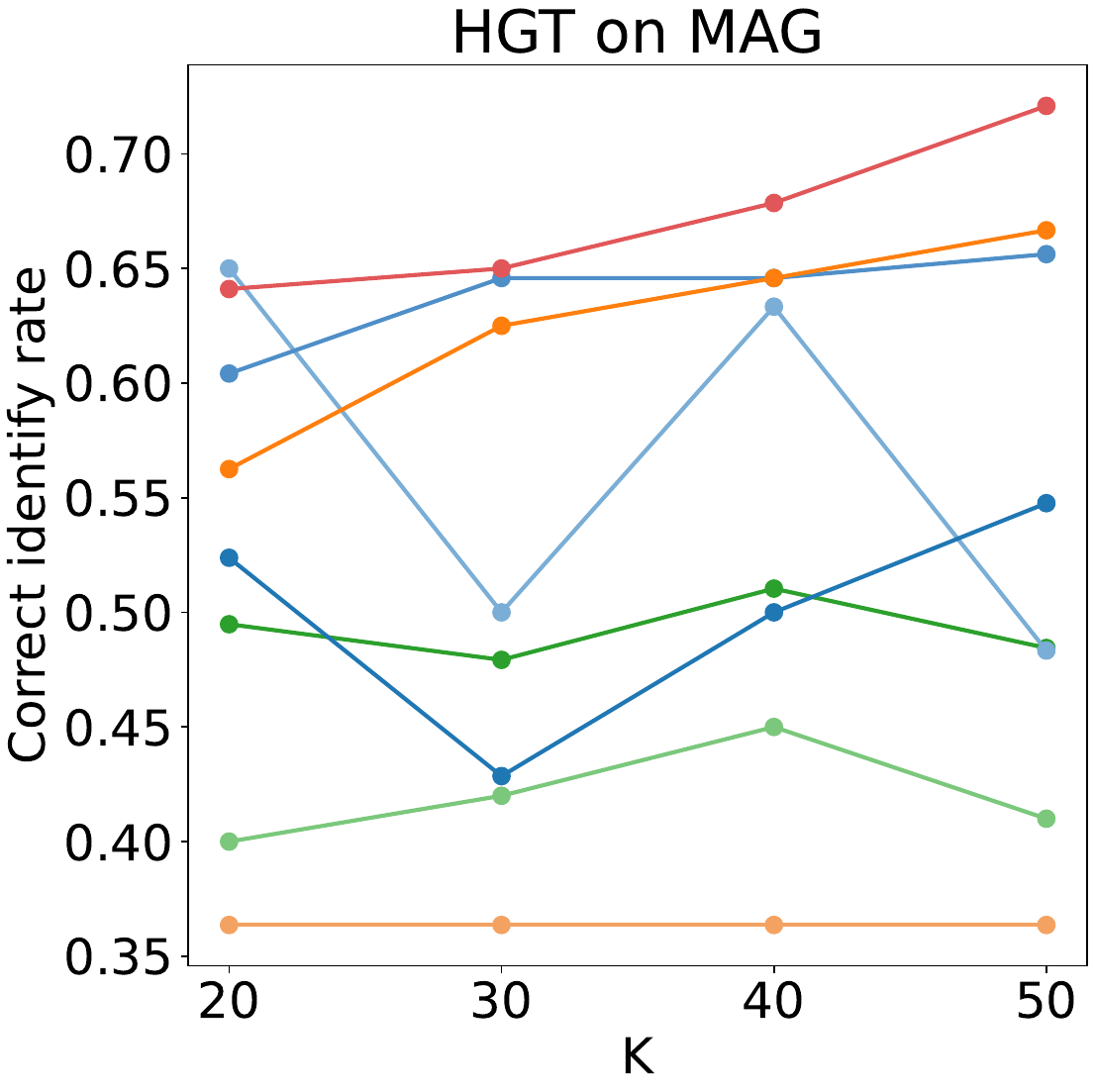}
    \end{minipage}
}

\vspace{0.5em}

\begin{minipage}{\linewidth}
\centering
\includegraphics[width=1\linewidth]{pdf/Li8.pdf}
\end{minipage}

\caption{Discrimination success rates of different explainers in correctly identifying the better-performing HGNN model under varying values of $K$. 
Results are reported on HAN, MAGNN, HetSANN, HGT with the ACM and MAG datasets. 
Higher values indicate stronger reliability of the explainer in distinguishing model quality.
The complete results across all model--dataset settings are deferred to Appendix.}
\label{Choose better model}
\end{figure*}

High-performance models should rely less on noise features, and an effective explainer should reveal this difference. To evaluate usability, we construct two categories of HGNN classifiers under the same data split and noise injection settings: ``good models'' with a test accuracy of $85\%\sim100\%$ and ``bad models'' with a test accuracy of $65\%\sim75\%$. Each explainer generates top-$K$ ($K=20,30,40,50$) feature explanations for $B=10$ key nodes. If the number of noise features for the ``good model'' is less than that for the ``bad model,'' the trial is considered successful. This process is repeated 200 times, and the discrimination success rates are plotted in Fig.~\ref{Choose better model}.

HiSE exhibits the strongest and most stable usability, consistently outperforming all the baselines across most settings. HGExplainer, GNNExplainer, and HENCE-X achieve limited and inconsistent performance. ZORRO achieves success rates only slightly above random, indicating weak discrimination ability. The performance of GraphLIME is the worst, with success rates approaching or falling below random in some scenarios.

HiSE's advantage lies in its hierarchical semantic explanation paradigm, which involves the construction of surrogate models within each single-semantic subgraph to approximate the original model’s decision boundary without disrupting the input feature distribution. This design enables effective suppression of noise interference and accurately reveals the true decision-making mechanisms of models under different semantic structures, thereby stably distinguishing the noise dependence differences between ``good'' and ``bad'' models. In contrast, perturbation-based methods (HGExplainer, GNNExplainer, HENCE-X) tend to disrupt the original feature distribution and amplify the incidental influence of irrelevant features, making importance estimates sensitive to random fluctuations. Despite using local surrogate models, GraphLIME is easily misled by a single semantic in heterogeneous graphs and fails to capture the hierarchical semantic decision-making mechanism of HGNNs.

\subsection{\textbf{Experiment 4 Cross-Semantic Explanation Capability}: Can HiSE Accurately Recover the Cross-Semantic Importance Learned by HGNNs? }

In this section, the cross-semantic explanation capability of HiSE, a key innovation of our method, is evaluated. We inject randomly generated noise subgraphs of varying densities into the original datasets and train HGNN models with accuracy $\geq 85\%$. Two metrics are evaluated: (1) \textbf{Cosine similarity (CS)} between the cross-semantic weights computed by HiSE and those provided by the original meta-path-based HGNN, where a higher CS indicates a more accurate characterization of semantic importance; (2) \textbf{Noise weight (NW)} assigned to noise subgraphs by HiSE, where a lower NW indicates stronger suppression of irrelevant semantics. All the experiments are repeated 200 times, and the average results are reported in Table~\ref{Meta-path weight}. We select HAN and MAGNN as baselines, as they are the only methods that provide explicit meta-path semantic weights for comparison.

HiSE achieves high CS across all the settings, consistently exceeding 99\% on HAN, indicating strong consistency with the original model's weight distribution at the cross-semantic level. Moreover, NW remains below 5\% in most cases, demonstrating effective suppression of noise subgraphs.

This performance arises because HiSE measures the discrepancy between the model's output under each semantic and the overall output via KL divergence, directly recovering a semantic weight distribution that reflects the original model's cross-semantic fusion mechanism.

\begin{table*}[t]
\centering
\caption{Comparison of HiSE and HGNN Cross-Semantic Weights.
Rows correspond to different noise settings (number and density), and columns report results on different HGNNs (HAN, MAGNN) and datasets (ACM, MAG). 
``CS" denotes the cosine similarity with the original HGNN ($\uparrow$), and ``NW" denotes the total weight assigned to noise subgraphs ($\downarrow$). 
Higher CS reflects stronger consistency with the original model, while lower NW indicates better noise suppression.}
\label{Meta-path weight}
\setlength{\tabcolsep}{4pt}
\renewcommand{\arraystretch}{1.1}
\begin{tabular}{c c cc cc cc cc}
\toprule
\multirow{2}{*}[-3ex]{\textbf{\makecell{Number of \\ Added Noise Subgraph}}}
&
\multirow{2}{*}[-3ex]{\textbf{\makecell{Noise Subgraph \\ Density}}}
& \multicolumn{4}{c}{\textbf{HAN}} 
& \multicolumn{4}{c}{\textbf{MAGNN}} \\
\cmidrule(lr){3-6} \cmidrule(lr){7-10}
& 
& \multicolumn{2}{c}{\textbf{ACM}} 
& \multicolumn{2}{c}{\textbf{MAG}} 
& \multicolumn{2}{c}{\textbf{ACM}} 
& \multicolumn{2}{c}{\textbf{MAG}} \\
\cmidrule(lr){3-10}
& 
& \textbf{CS $\uparrow$}& \textbf{NW $\downarrow$}& \textbf{CS $\uparrow$}& \textbf{NW $\downarrow$}& \textbf{CS $\uparrow$}& \textbf{NW $\downarrow$}& \textbf{CS $\uparrow$}& \textbf{NW $\downarrow$}\\
\midrule

0 & /                 
& 99.76\% & /     
& 99.83\% & / & 91.64\% & /       
& 99.12\% & /\\

1 & 0.5\%             
& 99.83\% & 0.23\%     
& 99.37\% & 0.52\%& 93.25\%& 2.51\%& 89.96\%& 2.96\%\\

1 & 5\%               
& 99.76\% & 0.25\%     
& 99.37\% & 0.42\%& 93.38\%& 4.32\%& 95.05\%& 1.74\%\\

1 & 50\%              
& 99.42\% & 0.02\%     
& 99.31\% & 0.39\%& 94.41\%& 2.74\%& 90.93\%& 2.41\%\\

2 & 5\% and 1\%       
& 99.85\% & 0.47\%     
& 98.87\% & 1.41\%& 92.18\%& 7.56\%       
& 90.65\%& 4.02\%\\

2 & 5\% and 10\%      
& 99.87\% & 0.47\%     
& 98.99\% & 1.06\%& 94.01\%& 12.13\% & 90.98\%& 4.76\%\\

2 & 50\% and 10\%     
& 99.81\% & 0.56\%     
& 99.35\% & 2.21\%& 94.20\%& 6.33\%& 93.03\%& 3.84\%\\

\bottomrule
\end{tabular}

\vspace{2mm}
\end{table*}

\subsection{\textbf{Experiment 5 Computational Efficiency}: How Efficient is HiSE?}

We also evaluate the computational efficiency of each explanation method. All the experiments are conducted on a server equipped with 4 NVIDIA RTX 4090 GPUs (48 GB each) and 4-way Intel Xeon Gold 5120 CPUs (112 cores), with only 1 GPU utilized. For each method, we randomly select 50 nodes, repeat the process 10 times, and report the average explanation time per node. Owing to GPU memory constraints, GraphLIME is implemented on the CPU.

As shown in Table~\ref{tab:efficiency_main}, HiSE achieves the shortest running time across all the settings. Compared with HENCE-X, HiSE achieves approximately \textbf{2--3 orders of magnitude} speedup. HGExplainer and GNNExplainer are approximately one order of magnitude slower than HiSE, while GraphLIME and ZORRO incur the highest running times in certain scenarios.

HiSE's efficiency stems from its hierarchical semantic mechanism, which decomposes the explanation into independent per-semantic surrogate modeling and lightweight cross-semantic aggregation, avoiding complex iterative search or combinatorial optimization. This design enables HiSE to efficiently reflect the decision-making mechanism of the model with minimal computational overhead. In contrast, HENCE-X and ZORRO rely on iterative feature selection with repeated evaluation of candidate subsets. The bottleneck of GraphLIME lies in kernel matrix computation, which imposes high memory demands.

\begin{table*}[htbp]
\centering
\caption{Efficiency comparison of different explainers under various HGNN classifiers and datasets. 
The reported values denote the average runtime (in seconds) of each node. 
The best results are highlighted in bold, respectively.}
\label{tab:efficiency_main}
\setlength{\tabcolsep}{4pt}
\renewcommand{\arraystretch}{1.1}
\begin{tabular}{c c ccccccc}
\toprule
\multirow{3}{*}{\textbf{Classifier}}
& \multirow{3}{*}{\textbf{Dataset}}
& \multicolumn{6}{c}{\textbf{Explainer (Time $\downarrow$)}} \\

\cmidrule(lr){3-8}
& 
& \multicolumn{3}{c}{\textbf{Homogeneous}}
& \multicolumn{2}{c}{\textbf{Heterogeneous}}
& \multicolumn{1}{c}{\textbf{Ours}} \\

\cmidrule(lr){3-5} \cmidrule(lr){6-7} \cmidrule(lr){8-8}
& 
& GraphLIME & ZORRO & GNNExplainer
& HGExplainer & HENCE-X
& HiSE \\

\midrule

\multirow{2}{*}{HAN}
& ACM & 202.94 & 7.81 & 1.63 & 2.74 & 83.70 & \textbf{0.37} \\
& MAG     & 159.33 & 32.07 & 2.78 & 6.47 & 203.62 & \textbf{0.10} \\
\midrule

\multirow{2}{*}{MAGNN}
& ACM & 141.16 & 722.99 & 3.92 & 10.99 & 256.74 & \textbf{2.45} \\
& MAG     & 54.94 & 16.33 & 4.94 & 10.30 & 129.72 & \textbf{2.52} \\
\midrule

\multirow{2}{*}{HetSANN}
& ACM & 102.49 & 1352.75 & 12.01 & 23.15 & 416.66 & \textbf{2.76} \\
& MAG     & 63.40 & 10.38 & 5.88 & 11.82 & 213.38 & \textbf{0.42} \\
\midrule

\multirow{2}{*}{HGT}
& ACM & 92.39 & 2805.72 & 4.49 & 9.14 & 847.49 & \textbf{3.77} \\
& MAG     & 69.23 & 3407.38 & 4.36 & 8.95 & 367.70 & \textbf{0.74} \\
\bottomrule
\end{tabular}
\end{table*}

\section{Conclusion}

In this paper, the interpretability of HGNNs is investigated. Existing explanation methods fail to adequately characterize the multisemantic interactions and hierarchical structures inherent in HGNNs and often incur high computational costs. To address these issues, we propose HiSE, a hierarchical semantic explanation framework that involves the construction of local surrogate models at the semantic level and the evaluation of feature contributions at the cross-semantic level, thereby faithfully reflecting the decision-making mechanism of HGNNs. Extensive experiments demonstrate that compared with existing methods, the effectiveness and efficiency of HiSE are superior and that it can stably identify influential features even in noisy environments.

Future work may explore finer-grained or adaptive semantic partitioning schemes, integration with diverse HGNN architectures, and modeling of richer interlayer semantic interactions to enhance the cross-task applicability of this paradigm.

\bibliographystyle{plain}
\bibliography{references}

@misc{kipf2017semisupervisedclassificationgraphconvolutional,
      title={Semi-Supervised Classification with Graph Convolutional Networks}, 
      author={Thomas N. Kipf and Max Welling},
      year={2017},
      eprint={1609.02907},
      archivePrefix={arXiv},
      primaryClass={cs.LG},
      url={https://arxiv.org/abs/1609.02907}, 
}

@misc{veličković2018graphattentionnetworks,
      title={Graph Attention Networks}, 
      author={Petar Veličković and Guillem Cucurull and Arantxa Casanova and Adriana Romero and Pietro Liò and Yoshua Bengio},
      year={2018},
      eprint={1710.10903},
      archivePrefix={arXiv},
      primaryClass={stat.ML},
      url={https://arxiv.org/abs/1710.10903}, 
}

@inproceedings{HGNN,
author = {Zhang, Chuxu and Song, Dongjin and Huang, Chao and Swami, Ananthram and Chawla, Nitesh V.},
title = {Heterogeneous Graph Neural Network},
year = {2019},
isbn = {9781450362016},
publisher = {Association for Computing Machinery},
address = {New York, NY, USA},
url = {https://doi.org/10.1145/3292500.3330961},
doi = {10.1145/3292500.3330961},
booktitle = {Proceedings of the 25th ACM SIGKDD International Conference on Knowledge Discovery \& Data Mining},
pages = {793–803},
numpages = {11},
location = {Anchorage, AK, USA},
series = {KDD '19}
}

@misc{ying2019gnnexplainergeneratingexplanationsgraph,
      title={GNNExplainer: Generating Explanations for Graph Neural Networks}, 
      author={Rex Ying and Dylan Bourgeois and Jiaxuan You and Marinka Zitnik and Jure Leskovec},
      year={2019},
      eprint={1903.03894},
      archivePrefix={arXiv},
      primaryClass={cs.LG},
      url={https://arxiv.org/abs/1903.03894}, 
}

@misc{luo2020parameterizedexplainergraphneural,
      title={Parameterized Explainer for Graph Neural Network}, 
      author={Dongsheng Luo and Wei Cheng and Dongkuan Xu and Wenchao Yu and Bo Zong and Haifeng Chen and Xiang Zhang},
      year={2020},
      eprint={2011.04573},
      archivePrefix={arXiv},
      primaryClass={cs.LG},
      url={https://arxiv.org/abs/2011.04573}, 
}

@misc{yuan2021explainabilitygraphneuralnetworks,
      title={On Explainability of Graph Neural Networks via Subgraph Explorations}, 
      author={Hao Yuan and Haiyang Yu and Jie Wang and Kang Li and Shuiwang Ji},
      year={2021},
      eprint={2102.05152},
      archivePrefix={arXiv},
      primaryClass={cs.LG},
      url={https://arxiv.org/abs/2102.05152}, 
}

@misc{ribeiro2016whyitrustyou,
      title={"Why Should I Trust You?": Explaining the Predictions of Any Classifier}, 
      author={Marco Tulio Ribeiro and Sameer Singh and Carlos Guestrin},
      year={2016},
      eprint={1602.04938},
      archivePrefix={arXiv},
      primaryClass={cs.LG},
      url={https://arxiv.org/abs/1602.04938}, 
}

@misc{huang2020graphlimelocalinterpretablemodel,
      title={GraphLIME: Local Interpretable Model Explanations for Graph Neural Networks}, 
      author={Qiang Huang and Makoto Yamada and Yuan Tian and Dinesh Singh and Dawei Yin and Yi Chang},
      year={2020},
      eprint={2001.06216},
      archivePrefix={arXiv},
      primaryClass={cs.LG},
      url={https://arxiv.org/abs/2001.06216}, 
}

@misc{wang2021heterogeneousgraphattentionnetwork,
      title={Heterogeneous Graph Attention Network}, 
      author={Xiao Wang and Houye Ji and Chuan Shi and Bai Wang and Peng Cui and P. Yu and Yanfang Ye},
      year={2021},
      eprint={1903.07293},
      archivePrefix={arXiv},
      primaryClass={cs.SI},
      url={https://arxiv.org/abs/1903.07293}, 
}

@misc{schlichtkrull2017modelingrelationaldatagraph,
      title={Modeling Relational Data with Graph Convolutional Networks}, 
      author={Michael Schlichtkrull and Thomas N. Kipf and Peter Bloem and Rianne van den Berg and Ivan Titov and Max Welling},
      year={2017},
      eprint={1703.06103},
      archivePrefix={arXiv},
      primaryClass={stat.ML},
      url={https://arxiv.org/abs/1703.06103}, 
}

@misc{chen2023srhetgnnsessionbasedrecommendationheterogeneousgraph,
      title={SR-HetGNN:Session-based Recommendation with Heterogeneous Graph Neural Network}, 
      author={Jinpeng Chen and Haiyang Li and Xudong Zhang and Fan Zhang and Senzhang Wang and Kaimin Wei and Jiaqi Ji},
      year={2023},
      eprint={2108.05641},
      archivePrefix={arXiv},
      primaryClass={cs.IR},
      url={https://arxiv.org/abs/2108.05641}, 
}

@inproceedings{Fu_2020, series={WWW ’20},
   title={MAGNN: Metapath Aggregated Graph Neural Network for Heterogeneous Graph Embedding},
   url={http://dx.doi.org/10.1145/3366423.3380297},
   DOI={10.1145/3366423.3380297},
   booktitle={Proceedings of The Web Conference 2020},
   publisher={ACM},
   author={Fu, Xinyu and Zhang, Jiani and Meng, Ziqiao and King, Irwin},
   year={2020},
   month=apr, pages={2331–2341},
   collection={WWW ’20} }

@article{PathSim,
author = {Sun, Yizhou and Han, Jiawei and Yan, Xifeng and Yu, Philip S. and Wu, Tianyi},
title = {PathSim: meta path-based top-K similarity search in heterogeneous information networks},
year = {2011},
issue_date = {August 2011},
publisher = {VLDB Endowment},
volume = {4},
number = {11},
issn = {2150-8097},
url = {https://doi.org/10.14778/3402707.3402736},
doi = {10.14778/3402707.3402736},
journal = {Proc. VLDB Endow.},
month = aug,
pages = {992–1003},
numpages = {12}
}

@article{Dong2017metapath2vecSR,
  title={metapath2vec: Scalable Representation Learning for Heterogeneous Networks},
  author={Yuxiao Dong and N. Chawla and Ananthram Swami},
  journal={Proceedings of the 23rd ACM SIGKDD International Conference on Knowledge Discovery and Data Mining},
  year={2017},
  url={https://api.semanticscholar.org/CorpusID:3919301}
}

@article{Tibshirani1996RegressionSA,
  title={Regression Shrinkage and Selection via the Lasso},
  author={Robert Tibshirani},
  journal={Journal of the royal statistical society series b-methodological},
  year={1996},
  volume={58},
  pages={267-288},
  url={https://api.semanticscholar.org/CorpusID:16162039}
}

@article{Efron_2004,
   title={Least angle regression},
   volume={32},
   ISSN={0090-5364},
   url={http://dx.doi.org/10.1214/009053604000000067},
   DOI={10.1214/009053604000000067},
   number={2},
   journal={The Annals of Statistics},
   publisher={Institute of Mathematical Statistics},
   author={Efron, Bradley and Hastie, Trevor and Johnstone, Iain and Tibshirani, Robert},
   year={2004},
   month=apr }

@ARTICLE{TheGNN,
  author={Scarselli, Franco and Gori, Marco and Tsoi, Ah Chung and Hagenbuchner, Markus and Monfardini, Gabriele},
  journal={IEEE Transactions on Neural Networks}, 
  title={The Graph Neural Network Model}, 
  year={2009},
  volume={20},
  number={1},
  pages={61-80},
  doi={10.1109/TNN.2008.2005605}}

@article{Battaglia2018RelationalIB,
  title={Relational inductive biases, deep learning, and graph networks},
  author={Peter W. Battaglia and Jessica B. Hamrick and Victor Bapst and Alvaro Sanchez-Gonzalez and Vin{\'i}cius Flores Zambaldi and Mateusz Malinowski and Andrea Tacchetti and David Raposo and Adam Santoro and Ryan Faulkner and Çaglar G{\"u}lçehre and H. Francis Song and Andrew J. Ballard and Justin Gilmer and George E. Dahl and Ashish Vaswani and Kelsey R. Allen and Charlie Nash and Victoria Langston and Chris Dyer and Nicolas Manfred Otto Heess and Daan Wierstra and Pushmeet Kohli and Matthew M. Botvinick and Oriol Vinyals and Yujia Li and Razvan Pascanu},
  journal={ArXiv},
  year={2018},
  volume={abs/1806.01261},
  url={https://api.semanticscholar.org/CorpusID:46935302}
}

@ARTICLE{9046288,
  author={Wu, Zonghan and Pan, Shirui and Chen, Fengwen and Long, Guodong and Zhang, Chengqi and Yu, Philip S.},
  journal={IEEE Transactions on Neural Networks and Learning Systems}, 
  title={A Comprehensive Survey on Graph Neural Networks}, 
  year={2021},
  volume={32},
  number={1},
  pages={4-24},
  doi={10.1109/TNNLS.2020.2978386}}

@misc{yuan2022explainabilitygraphneuralnetworks,
      title={Explainability in Graph Neural Networks: A Taxonomic Survey}, 
      author={Hao Yuan and Haiyang Yu and Shurui Gui and Shuiwang Ji},
      year={2022},
      eprint={2012.15445},
      archivePrefix={arXiv},
      primaryClass={cs.LG},
      url={https://arxiv.org/abs/2012.15445}, 
}

@article{MOTIE2024122156,
title = {Financial fraud detection using graph neural networks: A systematic review},
journal = {Expert Systems with Applications},
volume = {240},
pages = {122156},
year = {2024},
issn = {0957-4174},
doi = {https://doi.org/10.1016/j.eswa.2023.122156},
url = {https://www.sciencedirect.com/science/article/pii/S0957417423026581},
author = {Soroor Motie and Bijan Raahemi}
}

@article{BARREDOARRIETA202082,
title = {Explainable Artificial Intelligence (XAI): Concepts, taxonomies, opportunities and challenges toward responsible AI},
journal = {Information Fusion},
volume = {58},
pages = {82-115},
year = {2020},
issn = {1566-2535},
doi = {https://doi.org/10.1016/j.inffus.2019.12.012},
url = {https://www.sciencedirect.com/science/article/pii/S1566253519308103},
author = {Alejandro {Barredo Arrieta} and Natalia Díaz-Rodríguez and Javier {Del Ser} and Adrien Bennetot and Siham Tabik and Alberto Barbado and Salvador Garcia and Sergio Gil-Lopez and Daniel Molina and Richard Benjamins and Raja Chatila and Francisco Herrera}
}

@inproceedings{10.1145/3534678.3539321,
author = {Hou, Zhenyu and Liu, Xiao and Cen, Yukuo and Dong, Yuxiao and Yang, Hongxia and Wang, Chunjie and Tang, Jie},
title = {GraphMAE: Self-Supervised Masked Graph Autoencoders},
year = {2022},
isbn = {9781450393850},
publisher = {Association for Computing Machinery},
address = {New York, NY, USA},
url = {https://doi.org/10.1145/3534678.3539321},
doi = {10.1145/3534678.3539321},
booktitle = {Proceedings of the 28th ACM SIGKDD Conference on Knowledge Discovery and Data Mining},
pages = {594–604},
numpages = {11},
location = {Washington DC, USA},
series = {KDD '22}
}

@ARTICLE{7536145,
  author={Shi, Chuan and Li, Yitong and Zhang, Jiawei and Sun, Yizhou and Yu, Philip S.},
  journal={IEEE Transactions on Knowledge and Data Engineering}, 
  title={A survey of heterogeneous information network analysis}, 
  year={2017},
  volume={29},
  number={1},
  pages={17-37},
  doi={10.1109/TKDE.2016.2598561}}

@article{10.25300/MISQ/2014/38.1.04,
author = {Martens, David and Provost, Foster},
title = {Explaining data-driven document classifications},
year = {2014},
issue_date = {March 2014},
publisher = {Society for Information Management and The Management Information Systems Research Center},
address = {USA},
volume = {38},
number = {1},
issn = {0276-7783},
url = {https://doi.org/10.25300/MISQ/2014/38.1.04},
doi = {10.25300/MISQ/2014/38.1.04},
journal = {MIS Q.},
month = mar,
pages = {73–100},
numpages = {28}
}

@misc{gilmer2017neuralmessagepassingquantum,
      title={Neural Message Passing for Quantum Chemistry}, 
      author={Justin Gilmer and Samuel S. Schoenholz and Patrick F. Riley and Oriol Vinyals and George E. Dahl},
      year={2017},
      eprint={1704.01212},
      archivePrefix={arXiv},
      primaryClass={cs.LG},
      url={https://arxiv.org/abs/1704.01212}, 
}

@article{ZHOU202057,
title = {Graph neural networks: A review of methods and applications},
journal = {AI Open},
volume = {1},
pages = {57-81},
year = {2020},
issn = {2666-6510},
doi = {https://doi.org/10.1016/j.aiopen.2021.01.001},
url = {https://www.sciencedirect.com/science/article/pii/S2666651021000012},
author = {Jie Zhou and Ganqu Cui and Shengding Hu and Zhengyan Zhang and Cheng Yang and Zhiyuan Liu and Lifeng Wang and Changcheng Li and Maosong Sun}
}

@inproceedings{10.1145/3366423.3380027,
author = {Hu, Ziniu and Dong, Yuxiao and Wang, Kuansan and Sun, Yizhou},
title = {Heterogeneous Graph Transformer},
year = {2020},
isbn = {9781450370233},
publisher = {Association for Computing Machinery},
address = {New York, NY, USA},
url = {https://doi.org/10.1145/3366423.3380027},
doi = {10.1145/3366423.3380027},
booktitle = {Proceedings of The Web Conference 2020},
pages = {2704–2710},
numpages = {7},
location = {Taipei, Taiwan},
series = {WWW '20}
}

@misc{wang2020surveyheterogeneousgraphembedding,
      title={A Survey on Heterogeneous Graph Embedding: Methods, Techniques, Applications and Sources}, 
      author={Xiao Wang and Deyu Bo and Chuan Shi and Shaohua Fan and Yanfang Ye and Philip S. Yu},
      year={2020},
      eprint={2011.14867},
      archivePrefix={arXiv},
      primaryClass={cs.SI},
      url={https://arxiv.org/abs/2011.14867}, 
}

@article{article,
author = {Bing, Rui and Yuan, Guan and Zhu, Mu and Meng, Fanrong and Ma, Huifang and Qiao, Shaojie},
year = {2022},
month = {12},
pages = {1-40},
title = {Heterogeneous graph neural networks analysis: a survey of techniques, evaluations and applications},
volume = {56},
journal = {Artificial Intelligence Review},
doi = {10.1007/s10462-022-10375-2}
}

@misc{li2023surveyexplainablegraphneural,
      title={A Survey of Explainable Graph Neural Networks: Taxonomy and Evaluation Metrics}, 
      author={Yiqiao Li and Jianlong Zhou and Sunny Verma and Fang Chen},
      year={2023},
      eprint={2207.12599},
      archivePrefix={arXiv},
      primaryClass={cs.LG},
      url={https://arxiv.org/abs/2207.12599}, 
}

@misc{cui2022interpretablegraphneuralnetworks,
      title={Interpretable Graph Neural Networks for Connectome-Based Brain Disorder Analysis}, 
      author={Hejie Cui and Wei Dai and Yanqiao Zhu and Xiaoxiao Li and Lifang He and Carl Yang},
      year={2022},
      eprint={2207.00813},
      archivePrefix={arXiv},
      primaryClass={q-bio.NC},
      url={https://arxiv.org/abs/2207.00813}, 
}

@article{OSSBOLL2024104616,
title = {Graph neural networks for clinical risk prediction based on electronic health records: A survey},
journal = {Journal of Biomedical Informatics},
volume = {151},
pages = {104616},
year = {2024},
issn = {1532-0464},
doi = {https://doi.org/10.1016/j.jbi.2024.104616},
url = {https://www.sciencedirect.com/science/article/pii/S1532046424000340},
author = {Heloísa {Oss Boll} and Ali Amirahmadi and Mirfarid Musavian Ghazani and Wagner Ourique de Morais and Edison Pignaton de Freitas and Amira Soliman and Farzaneh Etminani and Stefan Byttner and Mariana Recamonde-Mendoza}
}

@misc{hong2019attentionbasedgraphneuralnetwork,
      title={An Attention-based Graph Neural Network for Heterogeneous Structural Learning}, 
      author={Huiting Hong and Hantao Guo and Yucheng Lin and Xiaoqing Yang and Zang Li and Jieping Ye},
      year={2019},
      eprint={1912.10832},
      archivePrefix={arXiv},
      primaryClass={cs.LG},
      url={https://arxiv.org/abs/1912.10832}, 
}

@misc{
funke2021hard,
title={Hard Masking for Explaining Graph Neural Networks},
author={Thorben Funke and Megha Khosla and Avishek Anand},
year={2021},
url={https://openreview.net/forum?id=uDN8pRAdsoC}
}

@article{Cliffs_delta, author = {Cliff, N.}, title = {Dominance statistics: Ordinal analyses to answer ordinal questions.}, journal = {Psychological Bulletin}, year = {1993}, volume = {114}, issue = {3}, pages = {494-509}, doi = {10.1037/0033-2909.114.3.494} }

@INPROCEEDINGS{HGExplainer,
  author={Mika, Grzegorz P. and Bouzeghoub, Amel and Węgrzyn-Wolska, Katarzyna and Neggaz, Yessin M.},
  booktitle={2023 IEEE/WIC International Conference on Web Intelligence and Intelligent Agent Technology (WI-IAT)}, 
  title={HGExplainer: Explainable Heterogeneous Graph Neural Network}, 
  year={2023},
  volume={},
  number={},
  pages={221-229},
  doi={10.1109/WI-IAT59888.2023.00035}}

@article{HENCEX,
author = {Lv, Ge and Zhang, Chen Jason and Chen, Lei},
title = {HENCE-X: Toward Heterogeneity-Agnostic Multi-Level Explainability for Deep Graph Networks},
year = {2023},
issue_date = {July 2023},
publisher = {VLDB Endowment},
volume = {16},
number = {11},
issn = {2150-8097},
url = {https://doi.org/10.14778/3611479.3611503},
doi = {10.14778/3611479.3611503},
journal = {Proc. VLDB Endow.},
month = jul,
pages = {2990–3003},
numpages = {14}
}

@misc{xPath,
      title={Towards Fine-Grained Explainability for Heterogeneous Graph Neural Network}, 
      author={Tong Li and Jiale Deng and Yanyan Shen and Luyu Qiu and Yongxiang Huang and Caleb Chen Cao},
      year={2023},
      eprint={2312.15237},
      archivePrefix={arXiv},
      primaryClass={cs.LG},
      url={https://arxiv.org/abs/2312.15237}, 
}

@inproceedings{HTGExplainer,
author = {Li, Jiazheng and Zhang, Chunhui and Zhang, Chuxu},
title = {Heterogeneous Temporal Graph Neural Network Explainer},
year = {2023},
isbn = {9798400701245},
publisher = {Association for Computing Machinery},
address = {New York, NY, USA},
url = {https://doi.org/10.1145/3583780.3614909},
doi = {10.1145/3583780.3614909},
booktitle = {Proceedings of the 32nd ACM International Conference on Information and Knowledge Management},
pages = {1298–1307},
numpages = {10},
location = {Birmingham, United Kingdom},
series = {CIKM '23}
}

@misc{ogbn_mag,
      title={Open Graph Benchmark: Datasets for Machine Learning on Graphs}, 
      author={Weihua Hu and Matthias Fey and Marinka Zitnik and Yuxiao Dong and Hongyu Ren and Bowen Liu and Michele Catasta and Jure Leskovec},
      year={2021},
      eprint={2005.00687},
      archivePrefix={arXiv},
      primaryClass={cs.LG},
      url={https://arxiv.org/abs/2005.00687}, 
}

@inproceedings{tail,
author = {Yang, Qiang and Ma, Changsheng and Zhang, Qiannan and Gao, Xin and Zhang, Chuxu and Zhang, Xiangliang},
title = {Interpretable Research Interest Shift Detection with Temporal Heterogeneous Graphs},
year = {2023},
isbn = {9781450394079},
publisher = {Association for Computing Machinery},
address = {New York, NY, USA},
url = {https://doi.org/10.1145/3539597.3570453},
doi = {10.1145/3539597.3570453},
booktitle = {Proceedings of the Sixteenth ACM International Conference on Web Search and Data Mining},
pages = {321–329},
numpages = {9},
location = {Singapore, Singapore},
series = {WSDM '23}
}

@misc{pagelink,
      title={PaGE-Link: Path-based Graph Neural Network Explanation for Heterogeneous Link Prediction}, 
      author={Shichang Zhang and Jiani Zhang and Xiang Song and Soji Adeshina and Da Zheng and Christos Faloutsos and Yizhou Sun},
      year={2023},
      eprint={2302.12465},
      archivePrefix={arXiv},
      primaryClass={cs.LG},
      url={https://arxiv.org/abs/2302.12465}, 
}

@article{LIU2024127274,
title = {A multi-level semantic-assisted unsupervised heterogeneous network representation learning model},
journal = {Neurocomputing},
volume = {574},
pages = {127274},
year = {2024},
issn = {0925-2312},
doi = {https://doi.org/10.1016/j.neucom.2024.127274},
url = {https://www.sciencedirect.com/science/article/pii/S0925231224000456},
author = {Qun Liu and Chengxin Peng and Shuyin Xia and Guoyin Wang}
}

@inproceedings{guan2025heterogeneous,
author = {Guan, Mingyu and Stokes, Jack W. and Luo, Qinlong and Liu, Fuchen and Mehta, Purvanshi and Nouri, Elnaz and Kim, Taesoo},
title = {Heterogeneous Graph Neural Network on Semantic Tree},
booktitle = {AAAI},
year = {2025},
month = {April},
url = {https://www.microsoft.com/en-us/research/publication/heterogeneous-graph-neural-network-on-semantic-tree/},
}

@article{10.1145/3568395,
author = {Guo, Naicheng and Liu, Xiaolei and Li, Shaoshuai and Ma, Qiongxu and Gao, Kaixin and Han, Bing and Zheng, Lin and Guo, Sheng and Guo, Xiaobo},
title = {Poincar\'{e} Heterogeneous Graph Neural Networks for Sequential Recommendation},
year = {2023},
issue_date = {July 2023},
publisher = {Association for Computing Machinery},
address = {New York, NY, USA},
volume = {41},
number = {3},
issn = {1046-8188},
url = {https://doi.org/10.1145/3568395},
doi = {10.1145/3568395},
journal = {ACM Trans. Inf. Syst.},
month = feb,
articleno = {63},
numpages = {26}
}

@ARTICLE{10478631,
  author={Van Belle, Rafaël and De Weerdt, Jochen},
  journal={IEEE Transactions on Knowledge and Data Engineering}, 
  title={SHINE: A Scalable Heterogeneous Inductive Graph Neural Network for Large Imbalanced Datasets}, 
  year={2024},
  volume={36},
  number={9},
  pages={4904-4915},
  doi={10.1109/TKDE.2024.3381240}}

@ARTICLE{10582518,
  author={Fang, Junfeng and Zhang, Guibin and Wang, Kun and Du, Wenjie and Duan, Yifan and Wu, Yuankai and Zimmermann, Roger and Chu, Xiaowen and Liang, Yuxuan},
  journal={IEEE Transactions on Knowledge and Data Engineering}, 
  title={On Regularization for Explaining Graph Neural Networks: An Information Theory Perspective}, 
  year={2024},
  volume={},
  number={},
  pages={1-14},
  doi={10.1109/TKDE.2024.3422328}}

@ARTICLE{10008205,
  author={Liu, Yanbei and Fan, Lianxi and Wang, Xiao and Xiao, Zhitao and Ma, Shuai and Pang, Yanwei and Lin, Jerry Chun-Wei},
  journal={IEEE Transactions on Neural Networks and Learning Systems}, 
  title={HGBER: Heterogeneous Graph Neural Network With Bidirectional Encoding Representation}, 
  year={2024},
  volume={35},
  number={7},
  pages={9340-9351},
  doi={10.1109/TNNLS.2022.3232709}}

@ARTICLE{10132398,
  author={Fang, Yueting and Wu, Hao and Zhao, Yiji and Zhang, Lei and Qin, Shaowei and Wang, Xin},
  journal={IEEE Transactions on Neural Networks and Learning Systems}, 
  title={Diversifying Collaborative Filtering via Graph Spreading Network and Selective Sampling}, 
  year={2024},
  volume={35},
  number={10},
  pages={13860-13873},
  doi={10.1109/TNNLS.2023.3272475}}

@ARTICLE{9925059,
  author={Xia, Lianghao and Huang, Chao and Xu, Yong and Dai, Peng and Bo, Liefeng},
  journal={IEEE Transactions on Neural Networks and Learning Systems}, 
  title={Multi-Behavior Graph Neural Networks for Recommender System}, 
  year={2024},
  volume={35},
  number={4},
  pages={5473-5487},
  doi={10.1109/TNNLS.2022.3204775}}

@ARTICLE{10059171,
  author={Lv, Qiujie and Chen, Guanxing and Yang, Ziduo and Zhong, Weihe and Chen, Calvin Yu-Chian},
  journal={IEEE Transactions on Neural Networks and Learning Systems}, 
  title={Meta Learning With Graph Attention Networks for Low-Data Drug Discovery}, 
  year={2024},
  volume={35},
  number={8},
  pages={11218-11230},
  doi={10.1109/TNNLS.2023.3250324}}

@ARTICLE{11204706,
  author={Jiang, Wenchao and Wu, Fangyue and Zhang, Fanlong and Chen, Quan and Zhao, Zhiming and Guo, Song},
  journal={IEEE Transactions on Neural Networks and Learning Systems}, 
  title={CRL-MM: Context-Aware Relational Learning and Multidimensional Matching for Few-Shot Knowledge Graph Completion}, 
  year={2026},
  volume={37},
  number={3},
  pages={1405-1419},
  doi={10.1109/TNNLS.2025.3615900}}

@ARTICLE{10115472,
  author={Gao, Chao and Yin, Shu and Wang, Haiqiang and Wang, Zhen and Du, Zhanwei and Li, Xuelong},
  journal={IEEE Transactions on Neural Networks and Learning Systems}, 
  title={Medical-Knowledge-Based Graph Neural Network for Medication Combination Prediction}, 
  year={2024},
  volume={35},
  number={10},
  pages={13246-13257},
  doi={10.1109/TNNLS.2023.3266490}}

@ARTICLE{9609099,
  author={Shehnepoor, Saeedreza and Togneri, Roberto and Liu, Wei and Bennamoun, Mohammed},
  journal={IEEE Transactions on Neural Networks and Learning Systems}, 
  title={HIN-RNN: A Graph Representation Learning Neural Network for Fraudster Group Detection With No Handcrafted Features}, 
  year={2023},
  volume={34},
  number={8},
  pages={4153-4166},
  doi={10.1109/TNNLS.2021.3123876}}

\clearpage
\appendix

\subsection{Pseudocode}

\subsubsection{HiSE Framework Overview}
\label{app:pseudocode}

Algorithm 1 summarizes the three-stage pipeline of HiSE. The heterogeneous graph is first decomposed into semantic subgraphs via meta-paths, followed by per-semantic sparse feature explanation through SSP, and finally unified across semantic views through CAI.

\begin{algorithm}
\caption{HiSE: Hierarchical Semantic Explanation for Heterogeneous Graph Neural Networks}
\label{alg:hise}
\begin{algorithmic}[1]
\Require{
    Heterogeneous graph $\mathcal{G} = (\mathcal{V}, \mathcal{E})$; 
    HGNN model $M$; 
    Target node $v_t$ to explain; 
    Pre-defined meta-paths $\mathcal{P} = \{\Phi_1, \Phi_2, \ldots, \Phi_m\}$; 
    Hop distance $K$; 
    Regularization strength $\lambda$
}
\Ensure{Semantic explanation $\mathbf{S} \in \mathbb{R}^{m \times n}$ ; Cross-semantic explanation $\mathbf{c} \in \mathbb{R}^n$}

\State \textbf{// Step 1: Semantic Graph Decomposition}
\State $\mathbb{G} \gets \text{Decompose}(\mathcal{G}, \mathcal{P})$ 

\State \textbf{// Step 2: Semantic Sparse Proxy}
\State $\mathbf{S} \gets \text{SSP}(\mathcal{G}, M, v_t, \mathbb{G}, K, \lambda)$ \Comment{Alg.~\ref{alg:ssp}}

\State \textbf{// Step 3: Cross-Semantic Aggregative Inversion}
\State $\mathbf{c} \gets \text{CAI}(\mathcal{G}, M, \mathbb{G}, \mathbf{S})$ \Comment{Alg.~\ref{alg:cai}}

\State \Return $\mathbf{S}$,  $\mathbf{c}$
\end{algorithmic}
\end{algorithm}

\subsubsection{Semantic Sparse Proxy}

Algorithm 2 details the per-semantic explanation procedure. For each semantic subgraph, SSP samples the K-hop neighborhood of the target node, and fits a weighted LASSO surrogate model to produce a semantic-level explanation.

\begin{algorithm}
\caption{Semantic Sparse Proxy (SSP)}
\label{alg:ssp}
\begin{algorithmic}[1]
\Require{
    Semantic subgraphs $\mathbb{G} = \{G_1, G_2, \ldots, G_m\}$; 
    HGNN model $M$; 
    Target node $v_t$ to explain; 
    Hop distance $K$; 
    Regularization strength $\lambda$
}
\Ensure{Semantic explanation matrix $\mathbf{S} \in \mathbb{R}^{m \times n}$}

\For{$k \gets 1$ to $m$}
    \State \textbf{// K-hop Weighted Neighborhood Sampling}
    \State $N_k \gets \{v_i \mid v_i \in V_k, \text{hop-distance}_{G_k}(v_t, v_i) \leq K\}$ 
    \State Extract edge weights $\eta_{uv}$ from $M$; if unavailable, set $\eta_{uv} \gets 1, \quad \forall (u,v) \in \mathcal{E}$

    \For{each $v_i \in N_k$}
        \State $\tilde{\mu}_i^{(k)} \gets \sum_{p \in \mathcal{P}_{\Phi_k}(v_t, v_i)} \prod_{(u,v)\in p} \eta_{uv}$ 
        \State $\mu_i^{(k)} \gets \frac{\tilde{\mu}_i^{(k)}}{\sum_{v_j \in N_k} \tilde{\mu}_j^{(k)}} \cdot |N_k|$ 
        \State $\bm{\hat{y}}^{\mathcal{G}}_i \gets M(\mathcal{G}, X_i)$ 
    \EndFor
    \State $\mathcal{D}_k \gets \{(X_i, \bm{\hat{y}}^{\mathcal{G}}_i, \mu_i^{(k)}) \mid v_i \in N_k\}$ 
    
    \State \textbf{// Weighted Sparse Proxy Fitting}
    \State $\bm{s}_{k,l} \gets \arg\min_{\bm{s}} \sum_{v_i \in N_k} \mu_i^{(k)} (\hat{y}^{\mathcal{G}}_{i,l} - \bm{s}^\top X_i)^2 + \lambda \|\bm{s}\|_1, \quad \forall l$ 
    \State $\bm{s}_k \gets \frac{1}{L} \sum_{l=1}^{L} \bm{s}_{k,l}$ 
\EndFor

\State \Return $\mathbf{S}$
\end{algorithmic}
\end{algorithm}

\subsubsection{Cross-Semantic Aggregative Inversion}

Algorithm 3 presents the cross-semantic explanation process, which quantifies contribution of each semantic view via KL divergence and aggregates per-semantic explanations into the cross-semantic-level explanation.

\begin{algorithm}
\caption{Cross-Semantic Aggregative Inversion (CAI)}
\label{alg:cai}
\begin{algorithmic}[1]
\Require{
    Heterogeneous graph $\mathcal{G} = (\mathcal{V}, \mathcal{E})$; 
    HGNN model $M$; 
    Semantic subgraphs $\mathbb{G} = \{G_1, G_2, \ldots, G_m\}$; 
    Semantic explanation matrix $\mathbf{S} \in \mathbb{R}^{m \times n}$
}
\Ensure{Cross-semantic explanation $\mathbf{c} \in \mathbb{R}^n$}

\For{$k \gets 1$ to $m$}
    \For{each $v_i \in \mathcal{N}$}
        \State $\bm{\hat{y}}^{G_k}_i \gets M(G_k, X_i)$ \Comment{Predict with single-semantic subgraph}
        \State $\bm{\hat{y}}^{\mathcal{G}}_i \gets M(\mathcal{G}, X_i)$ \Comment{Predict with full graph}
    \EndFor
    \State $\gamma_k \gets - \sum_{v_i \in \mathcal{N}} \bm{\hat{y}}^{\mathcal{G}}_i \log \frac{\bm{\hat{y}}^{\mathcal{G}}_i}{\bm{\hat{y}}^{G_k}_i}$ 
\EndFor

\State $w_k \gets \frac{\exp(\gamma_k)}{\sum_{k=1}^{m} \exp(\gamma_k)}, \quad k = 1, \ldots, m$ 

\State $\mathbf{c} \gets \sum_{k=1}^{m} w_k \cdot \bm{s}_k$ 

\State \Return $\mathbf{c}$
\end{algorithmic}
\end{algorithm}

\subsection{LassoLARS algorithm}

The LassoLARS algorithm initializes the parameters with a zero vector, i.e., $\bm{s}_{k,l}^{(0)}=\mathbf{0}$, and defines the initial residual as:
\begin{equation}
\bm{r}^{(0)}_l
=
\hat{\bm{y}}^{\mathcal{G}}_{:,l}
-
\mathbf{0}.
\end{equation}

At the $t$-th iteration, the correlations between the weighted residuals and each feature are computed:
\begin{equation}
Correlation_{j,l}^{(t)}
=
\left|
\sum_{v_i \in N_k}
\mu_i^{(k)} \, X_{ij} \, r_{i,l}^{(t)}
\right|,
\quad j=1,\ldots,n,
\end{equation}
and the feature with the largest absolute correlation is selected:
\begin{equation}
j_l^\star
=
\arg\max_j
Correlation_{j,l}^{(t)} ,
\end{equation}
which is then added to the active set $Q_l$.

Subsequently, the algorithm updates the model parameters along the equiangular direction $\boldsymbol{\theta}_{Q_l}$ within the current active feature subspace:
\begin{equation}
\boldsymbol{\theta}_{Q_l}
=
X_{:,Q_l}
\left(
X_{:,Q_l}^\top
\operatorname{diag}(\bm{\mu^{(k)}})
X_{:,Q_l}
\right)^{-1}
\mathbf{1},
\end{equation}
where $\operatorname{diag}(\bm{\mu^{(k)}})$ is the diagonal matrix constructed from the weight vector $\bm{\mu^{(k)}}=(\mu_1^{(k)},\ldots,\mu_{|N_k|}^{(k)})$.

The parameter and residual updates take the form:
\begin{equation}
\bm{s}_{k,l}^{(t+1)}
=
\bm{s}_{k,l}^{(t)}
+
\alpha_l \, \boldsymbol{\theta}_{Q_l},
\end{equation}
\begin{equation}
\bm{r}_l^{(t+1)}
=
\hat{\bm{y}}^{\mathcal{G}}_{:,l}
-
X \bm{s}_{k,l}^{(t+1)} .
\end{equation}

The step size $\alpha_l$ is chosen such that:
\begin{equation}
Correlation_{j,l}^{(t+1)}
=
Correlation_{j_l^\star,l}^{(t+1)},
\end{equation}
i.e., when the weighted correlation of an inactive feature first reaches the absolute correlation of the current active features, it is added to the active set $Q_l$, and joint updating continues within the active subspace.

Upon convergence, the final semantic-level feature explanation $
\bm{s}_{k,l}^{(\text{final})}$ for class $l$ under semantic $\Phi_k$ is obtained.


\section{Dataset Details}
\begin{table*}[t]
\centering
\caption{Statistics of the heterogeneous graph datasets used in our experiments. 
We report relation types, numbers of nodes and edges, feature dimensionality, data splits, and meta-paths. 
\textit{Density} (in parentheses) denotes the edge density of the single-semantic subgraph induced by each meta-path.}
\label{tab:dataset_statistics}
\setlength{\tabcolsep}{4pt}
\renewcommand{\arraystretch}{1.1}

\begin{tabular}{c c c c c c c c c c}
\toprule
\textbf{Dataset} 
& \textbf{Relation (A-B)} 
& \textbf{\#Nodes(A)} 
& \textbf{\#Nodes(B)} 
& \textbf{\#Edges} 
& \textbf{Feature Dim} 
& \textbf{Train} 
& \textbf{Val} 
& \textbf{Test} 
& \textbf{Meta-Path (Density)} \\
\midrule

\multirow{2}{*}{ACM}
& Paper--Author  & 4025 & 17431 & 13407 & \multirow{2}{*}{1870} & \multirow{2}{*}{808} & \multirow{2}{*}{401} & \multirow{2}{*}{2816} & \textit{PAP (0.36\%)} \\
& Paper--Subject & 4025 & 73    & 4025  &       &       &       &       & \textit{PSP (26.86\%)} \\
\midrule

\multirow{2}{*}{MAG}
& Paper--Author & 5000 & 25267 & 32683 & \multirow{2}{*}{128} & \multirow{2}{*}{4000} & \multirow{2}{*}{500} & \multirow{2}{*}{500} & \textit{PAP (0.15\%)} \\
& Paper--Field  & 5000 & 5449  & 50658 &       &       &       &       & \textit{PFP (99.99\%)} \\
\bottomrule
\end{tabular}
\end{table*}


\begin{figure*}[!htbp]
\centering
\subfloat[HAN on MAG]
{
    \begin{minipage}[b]{0.22\linewidth}
    \centering
    \includegraphics[width=\linewidth]{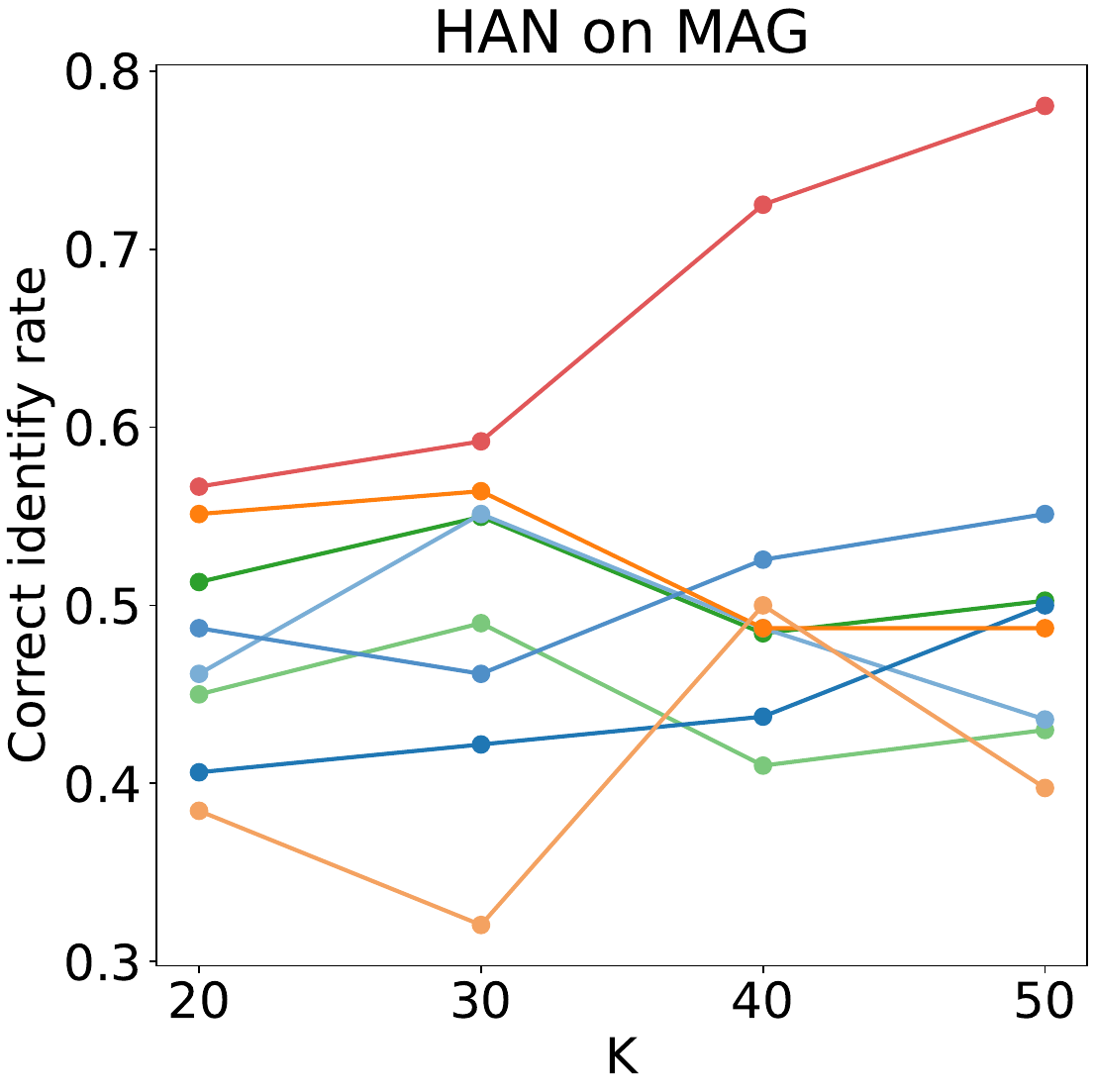}
    \end{minipage}
}
\subfloat[MAGNN on MAG]
{
    \begin{minipage}[b]{0.22\linewidth}
    \centering
    \includegraphics[width=\linewidth]{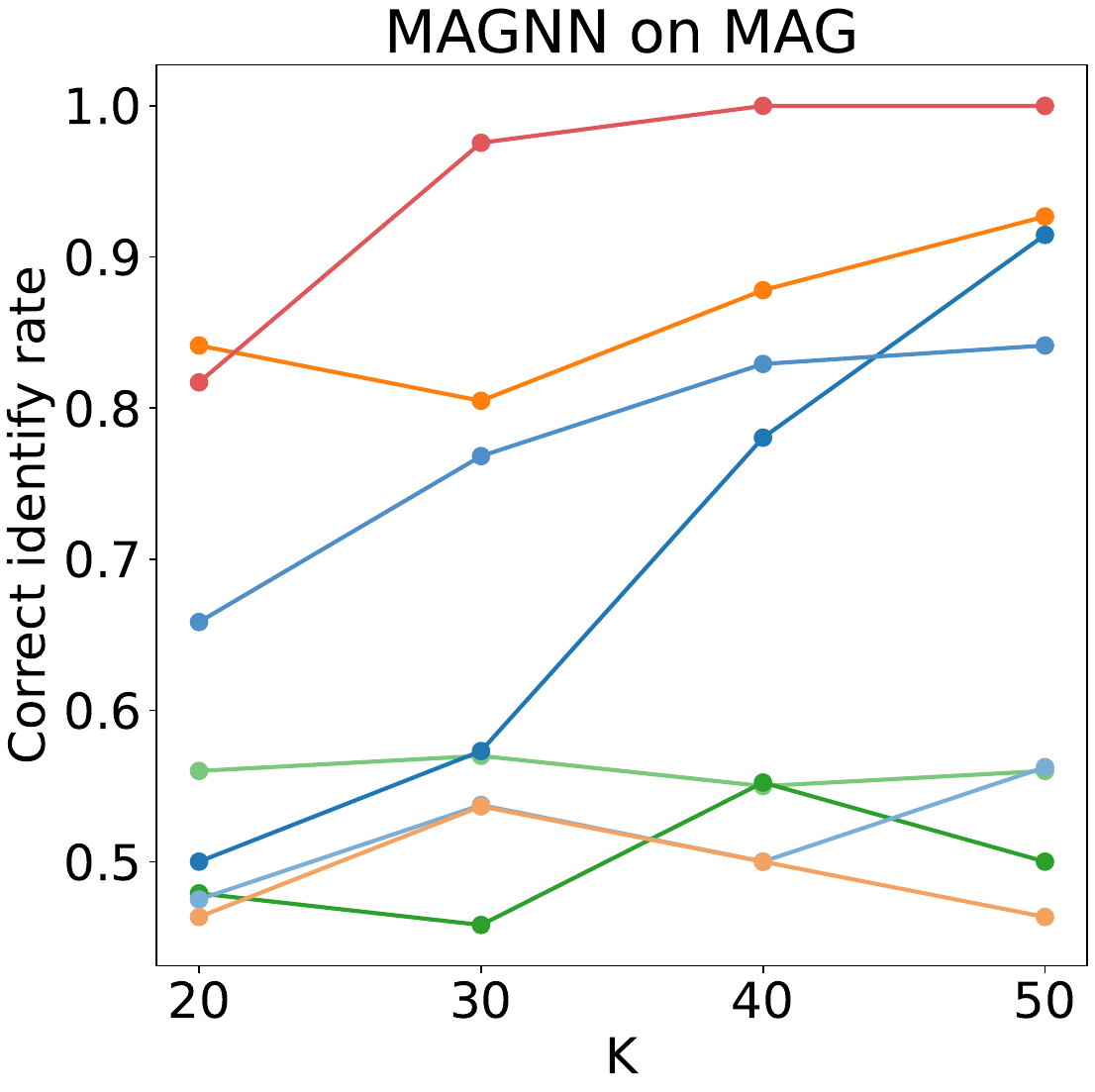}
    \end{minipage}
}
\subfloat[HetSANN on ACM]
{
    \begin{minipage}[b]{0.22\linewidth}
    \centering
    \includegraphics[width=\linewidth]{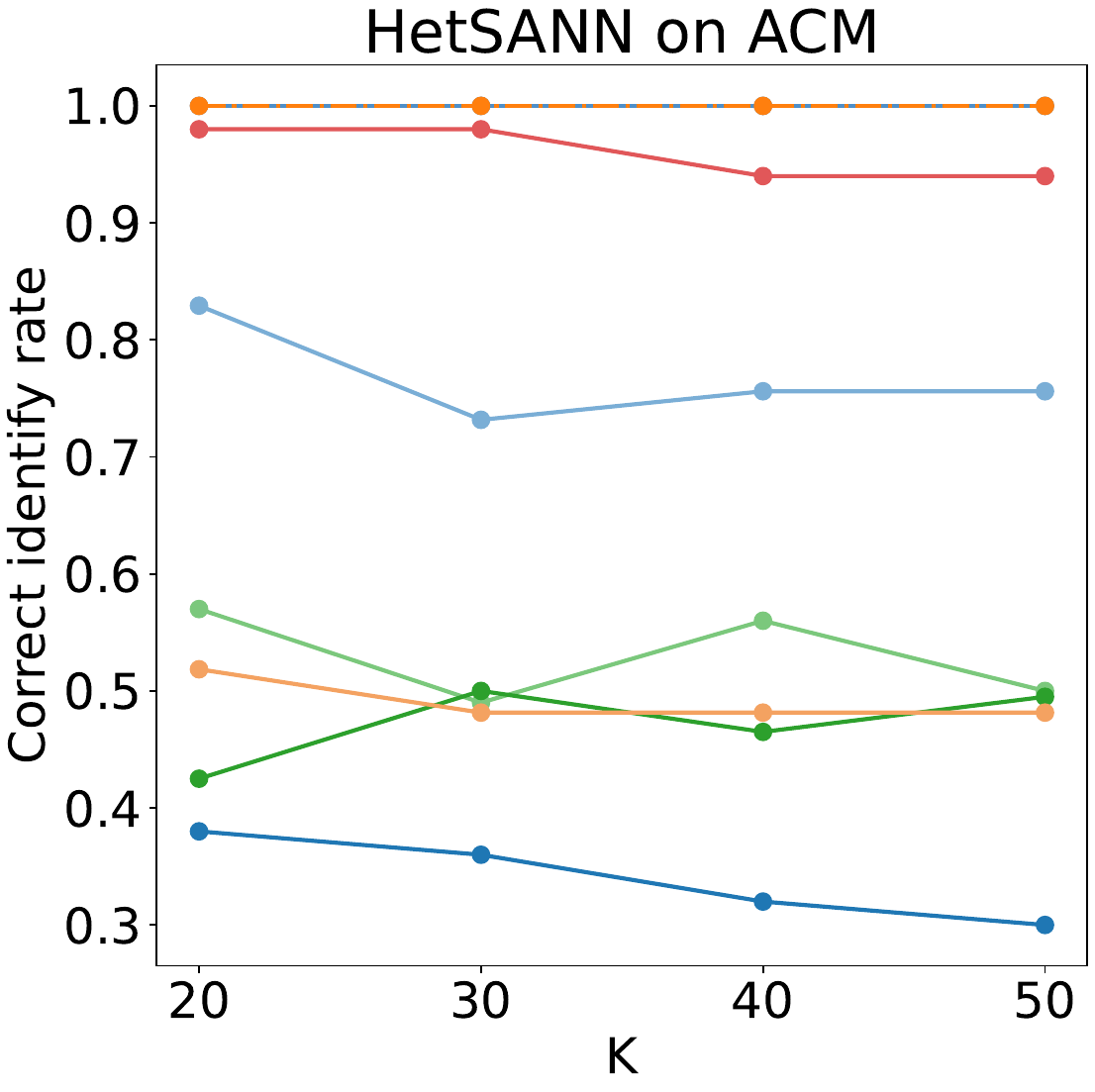}
    \end{minipage}
}
\subfloat[HGT on ACM]
{
    \begin{minipage}[b]{0.22\linewidth}
    \centering
    \includegraphics[width=\linewidth]{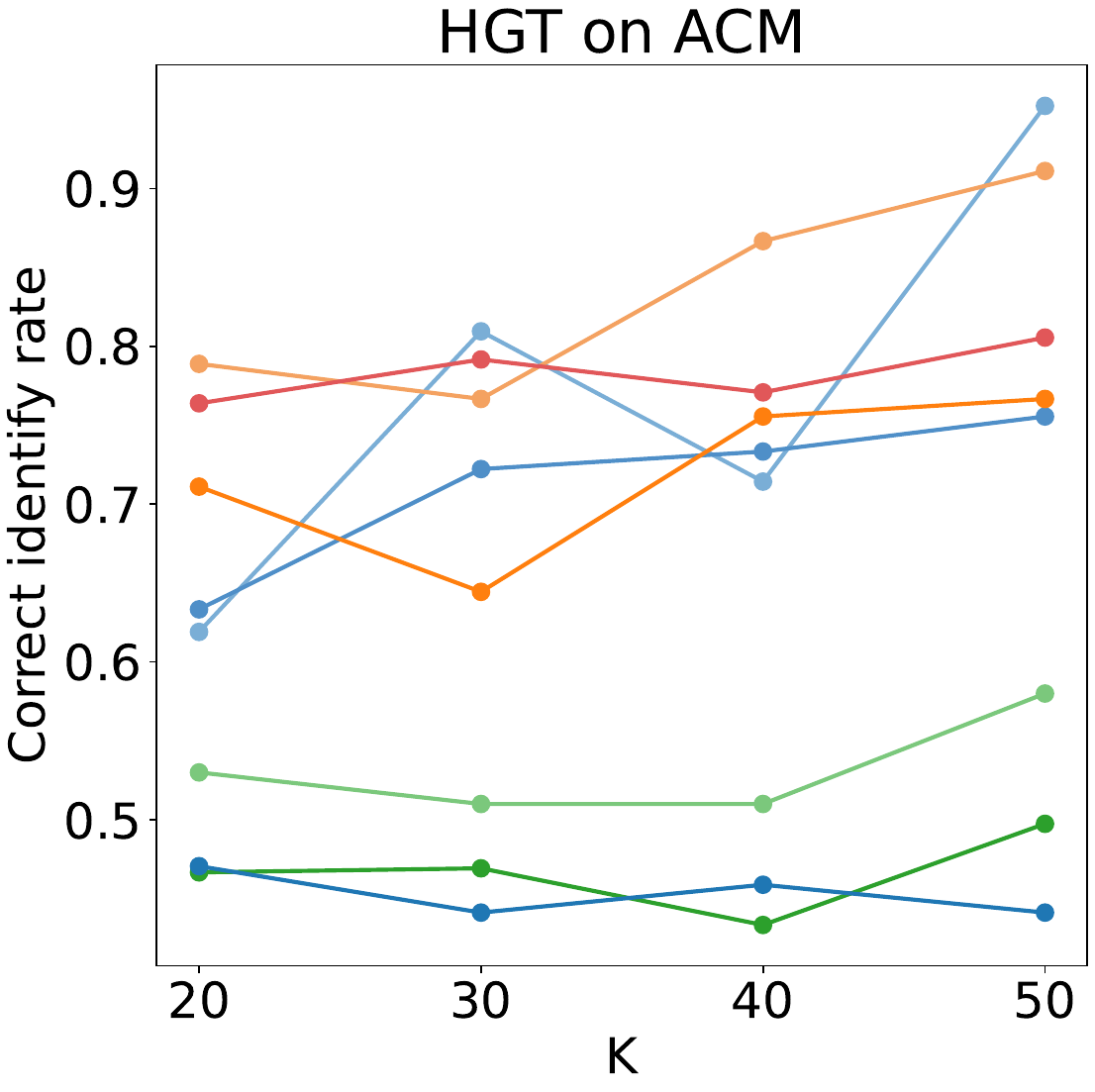}
    \end{minipage}
}

\vspace{0.5em}

\begin{minipage}{\linewidth}
\centering
\includegraphics[width=1\linewidth]{pdf/Li8.pdf}
\end{minipage}

\caption{\textbf{Usability Results}: Discrimination success rates of different explainers in correctly identifying the better-performing HGNN model under varying values of $K$. 
Results are reported on HAN, MAGNN, HetSANN, HGT with the ACM and MAG datasets. 
Higher values indicate stronger reliability of the explainer in distinguishing model quality.}
\label{app:Choose better model}
\end{figure*}

\begin{figure*}[!htbp]
\centering
\begin{minipage}{\linewidth}
\centering
\subfloat[HAN on ACM ($K\!=\!20$)]{
    \includegraphics[width=0.25\linewidth]{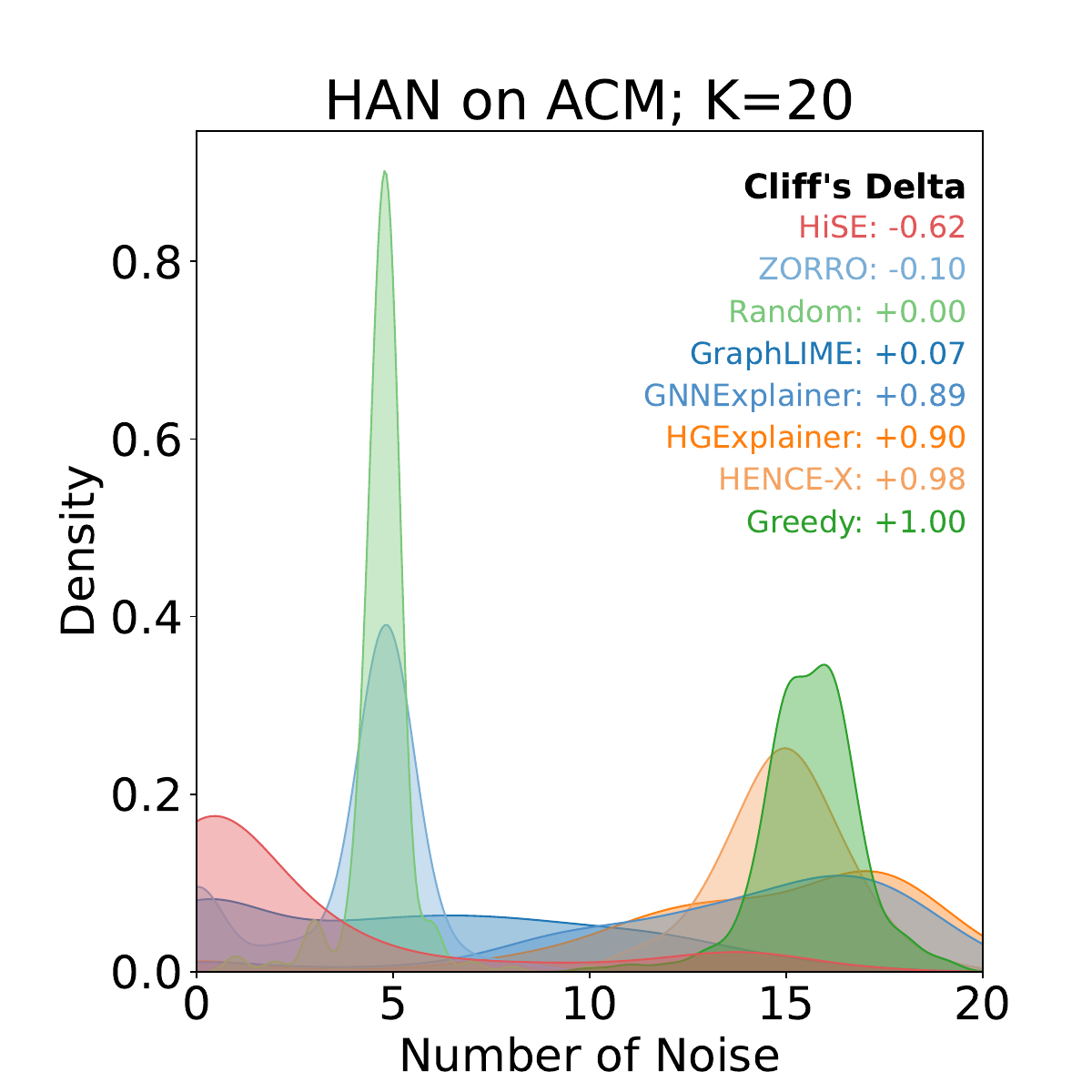}
}
\subfloat[HAN on MAG ($K\!=\!10$)]{
    \includegraphics[width=0.25\linewidth]{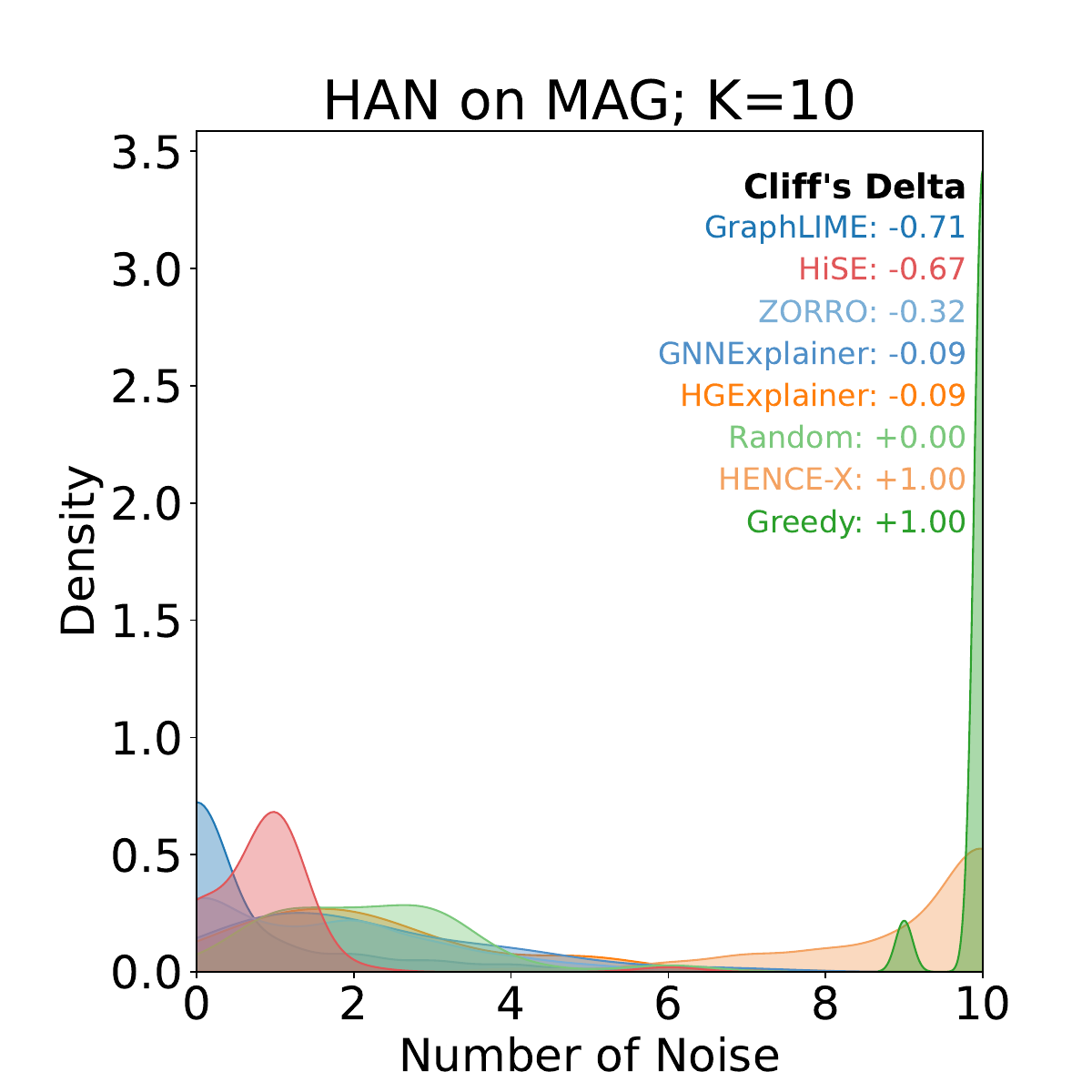}
}
\subfloat[HAN on MAG ($K\!=\!20$)]{
    \includegraphics[width=0.25\linewidth]{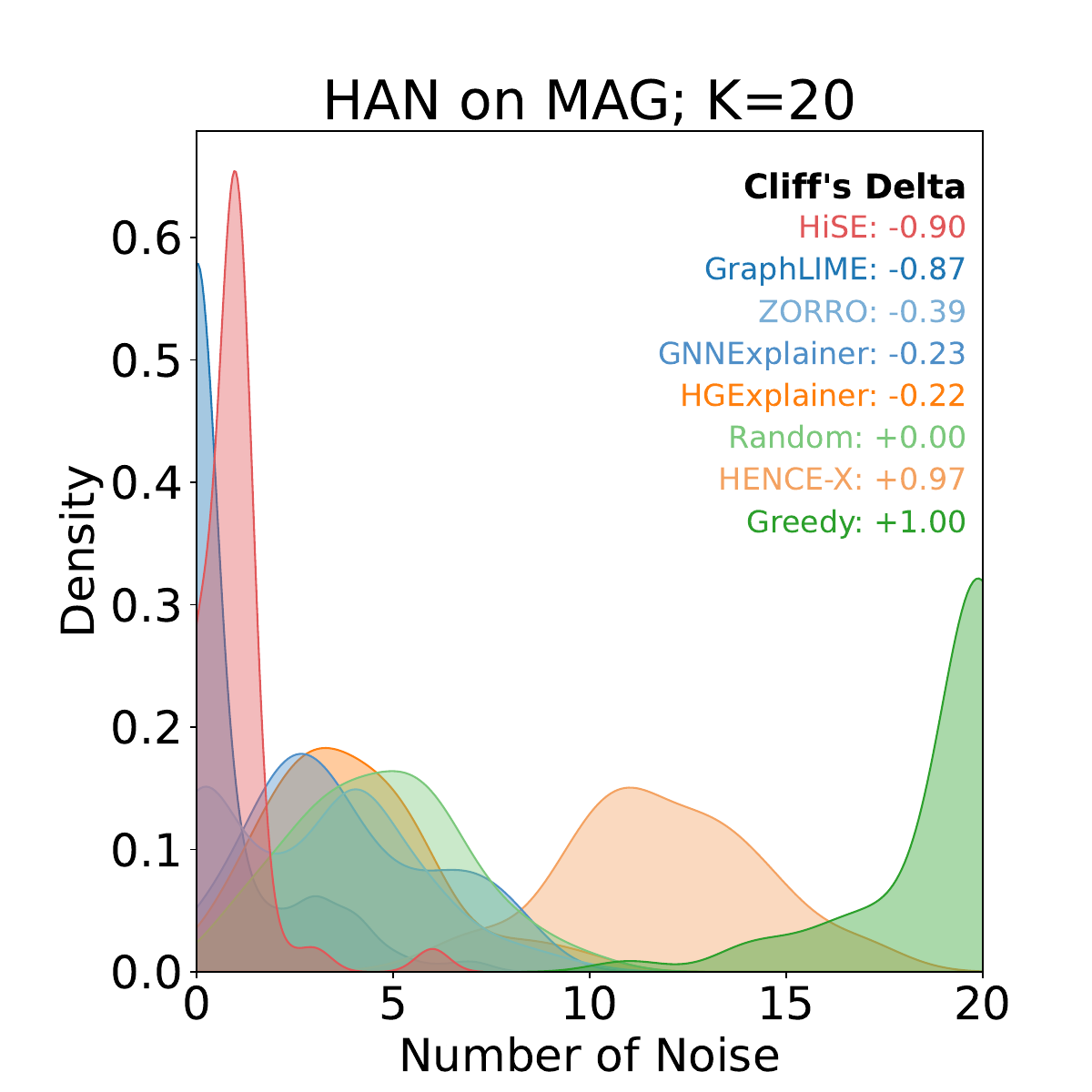}
}
\end{minipage}

\vspace{0.5em}

\begin{minipage}{\linewidth}
\centering
\subfloat[MAGNN on ACM ($K\!=\!10$)]{
    \includegraphics[width=0.25\linewidth]{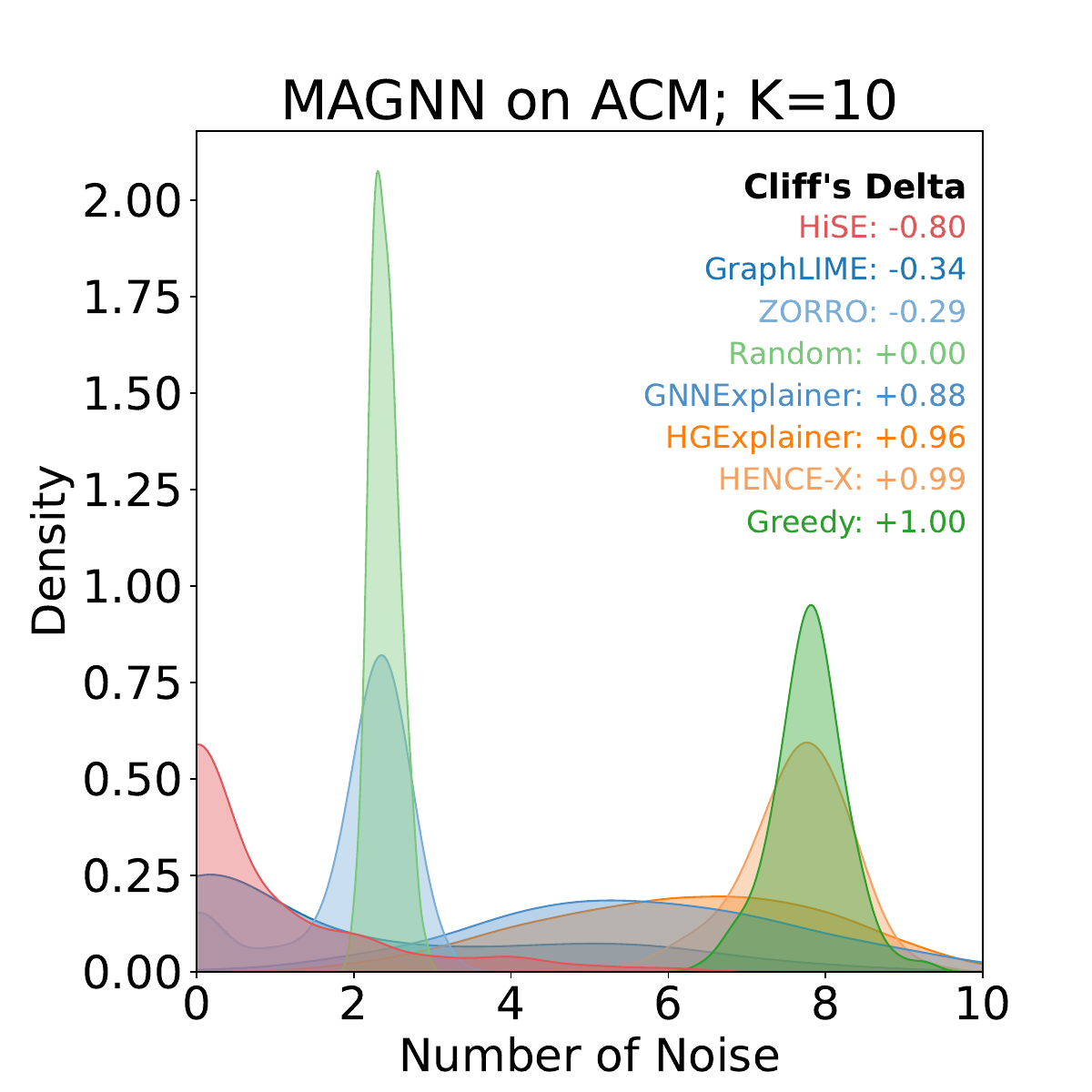}
}
\subfloat[MAGNN on MAG ($K\!=\!10$)]{
    \includegraphics[width=0.25\linewidth]{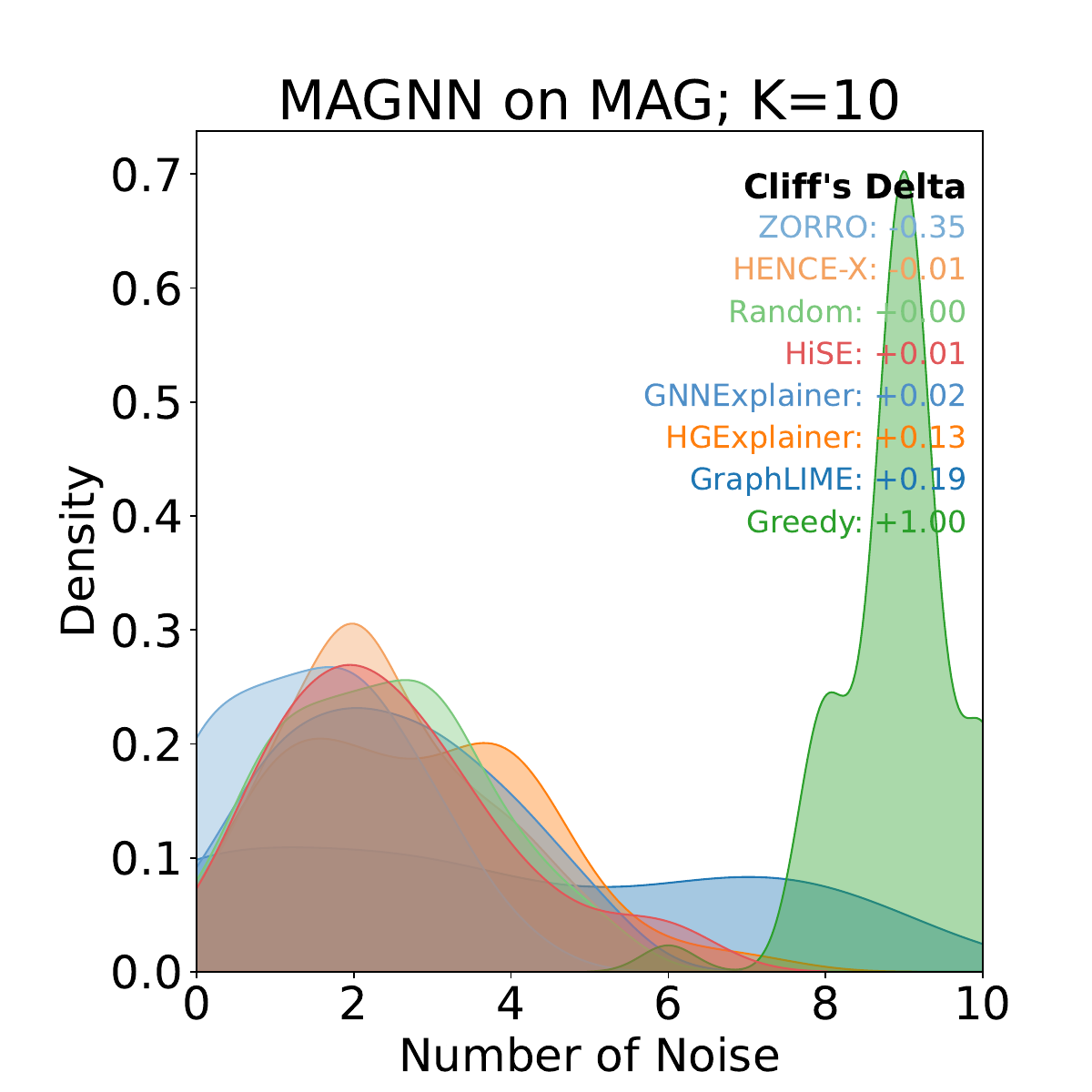}
}
\subfloat[MAGNN on MAG ($K\!=\!20$)]{
    \includegraphics[width=0.25\linewidth]{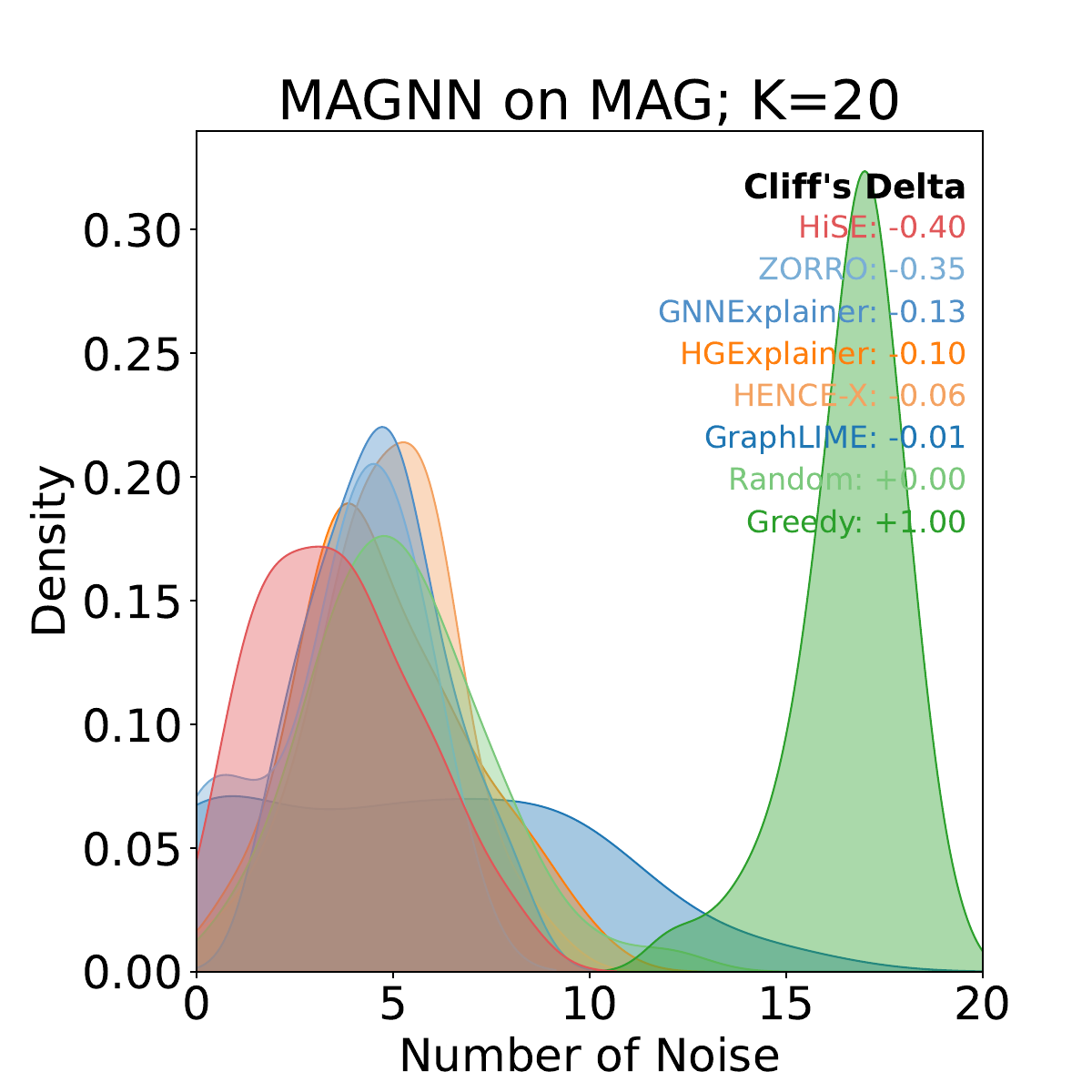}
}
\end{minipage}

\vspace{0.5em}

\begin{minipage}{\linewidth}
\centering
\subfloat[HetSANN on ACM ($K\!=\!10$)]{
    \includegraphics[width=0.25\linewidth]{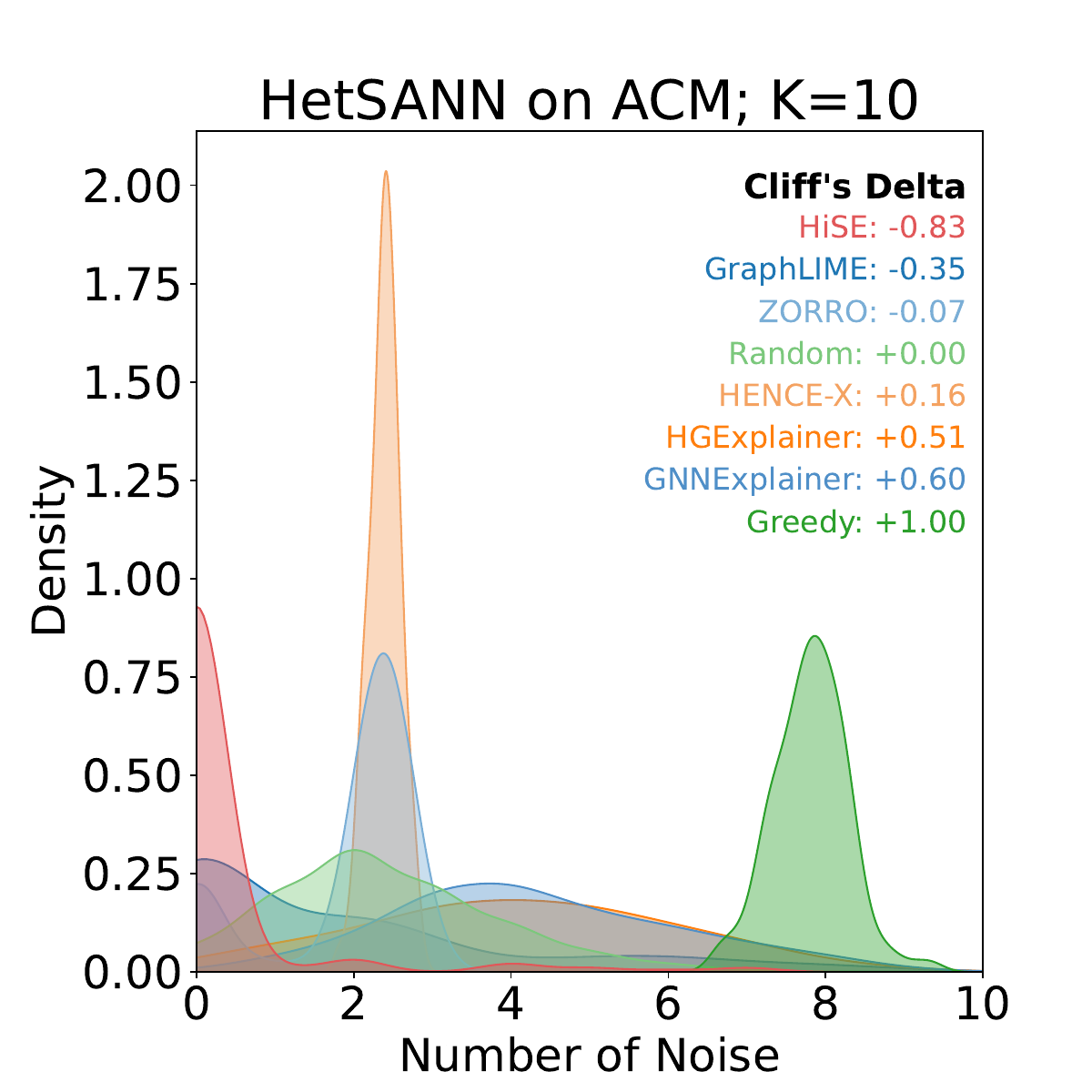}
}
\subfloat[HetSANN on ACM ($K\!=\!20$)]{
    \includegraphics[width=0.25\linewidth]{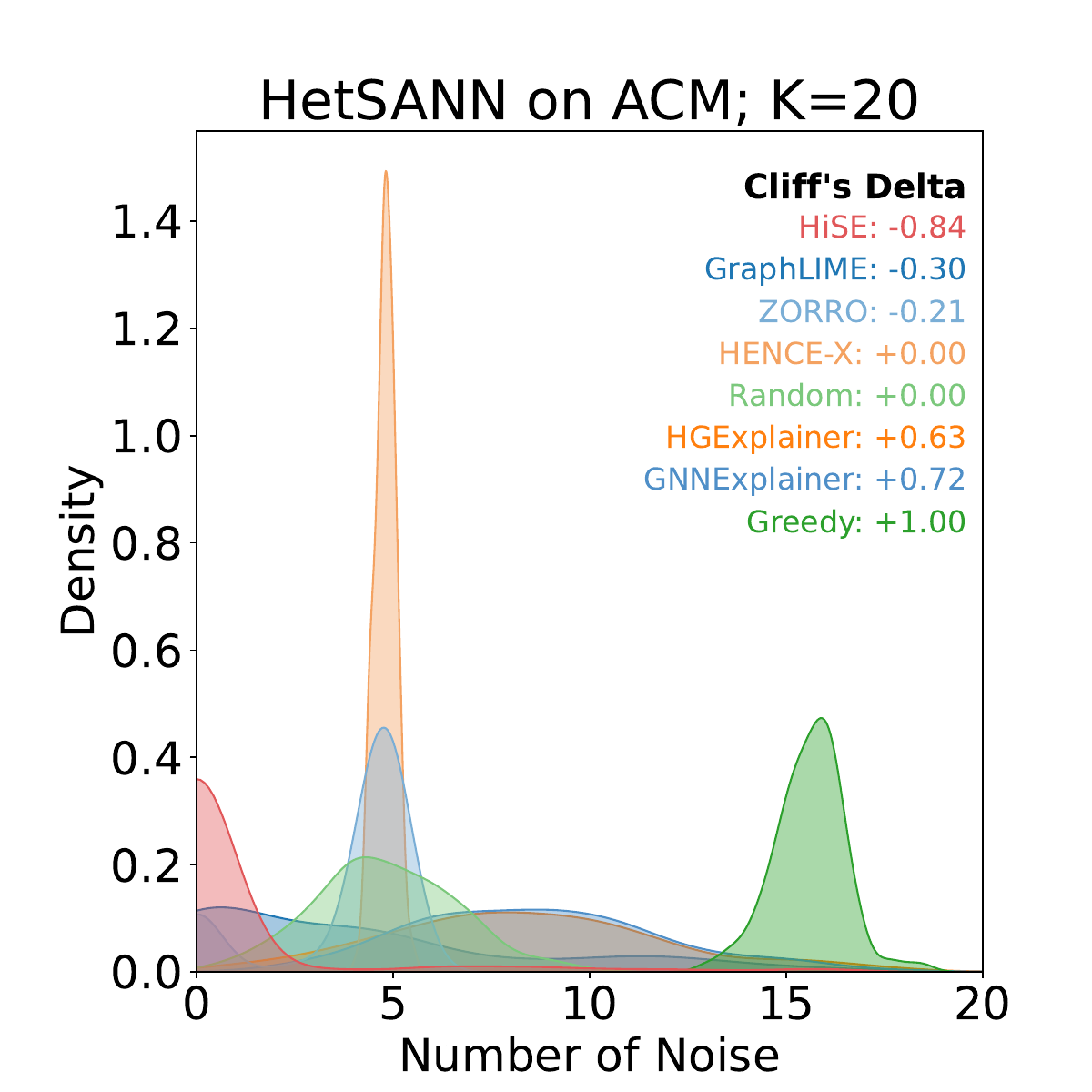}
}
\subfloat[HetSANN on MAG ($K\!=\!20$)]{
    \includegraphics[width=0.25\linewidth]{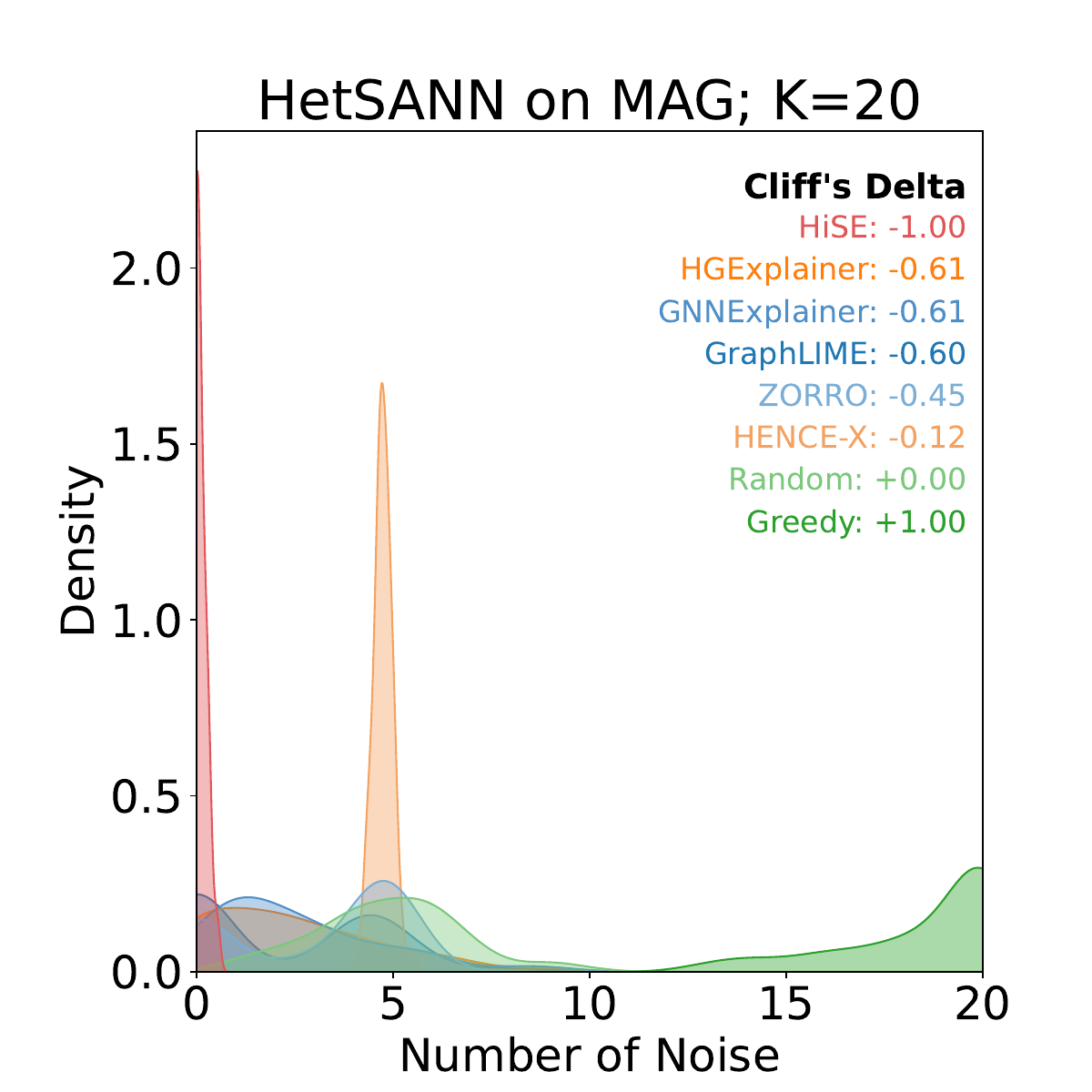}
}
\end{minipage}

\vspace{0.5em}

\begin{minipage}{\linewidth}
\centering
\subfloat[HGT on ACM ($K\!=\!10$)]{
    \includegraphics[width=0.25\linewidth]{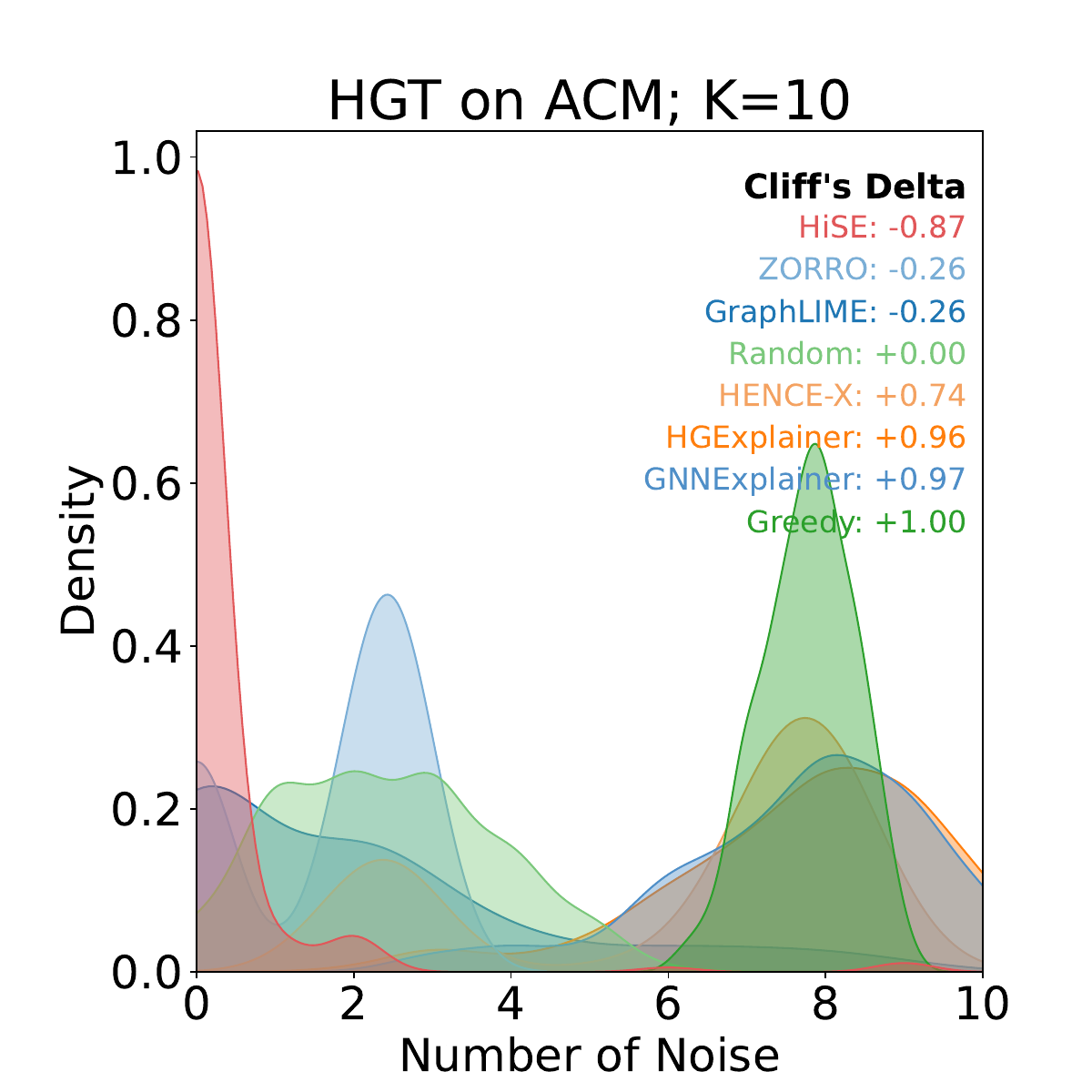}
}
\subfloat[HGT on ACM ($K\!=\!20$)]{
    \includegraphics[width=0.25\linewidth]{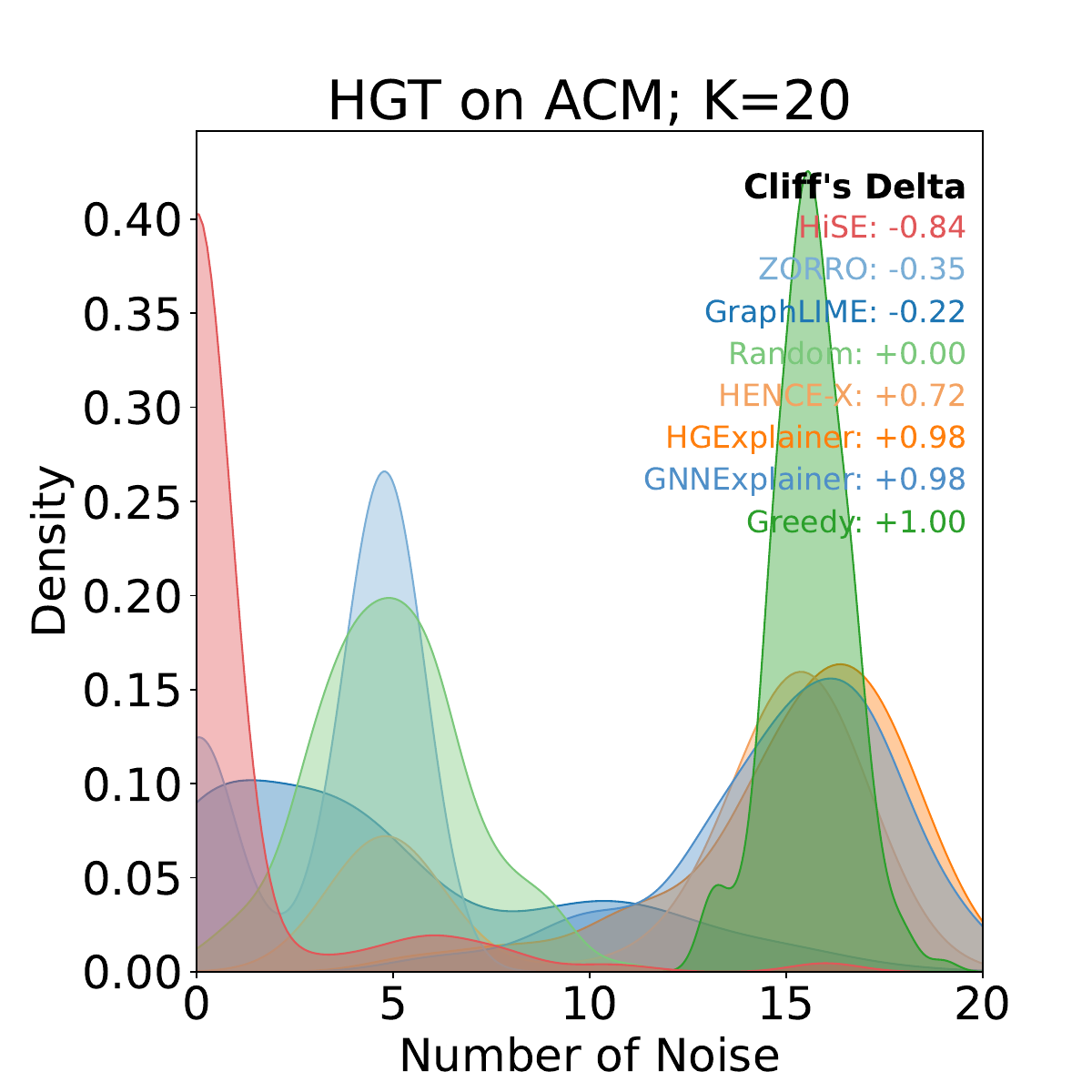}
}
\subfloat[HGT on MAG ($K\!=\!10$)]{
    \includegraphics[width=0.25\linewidth]{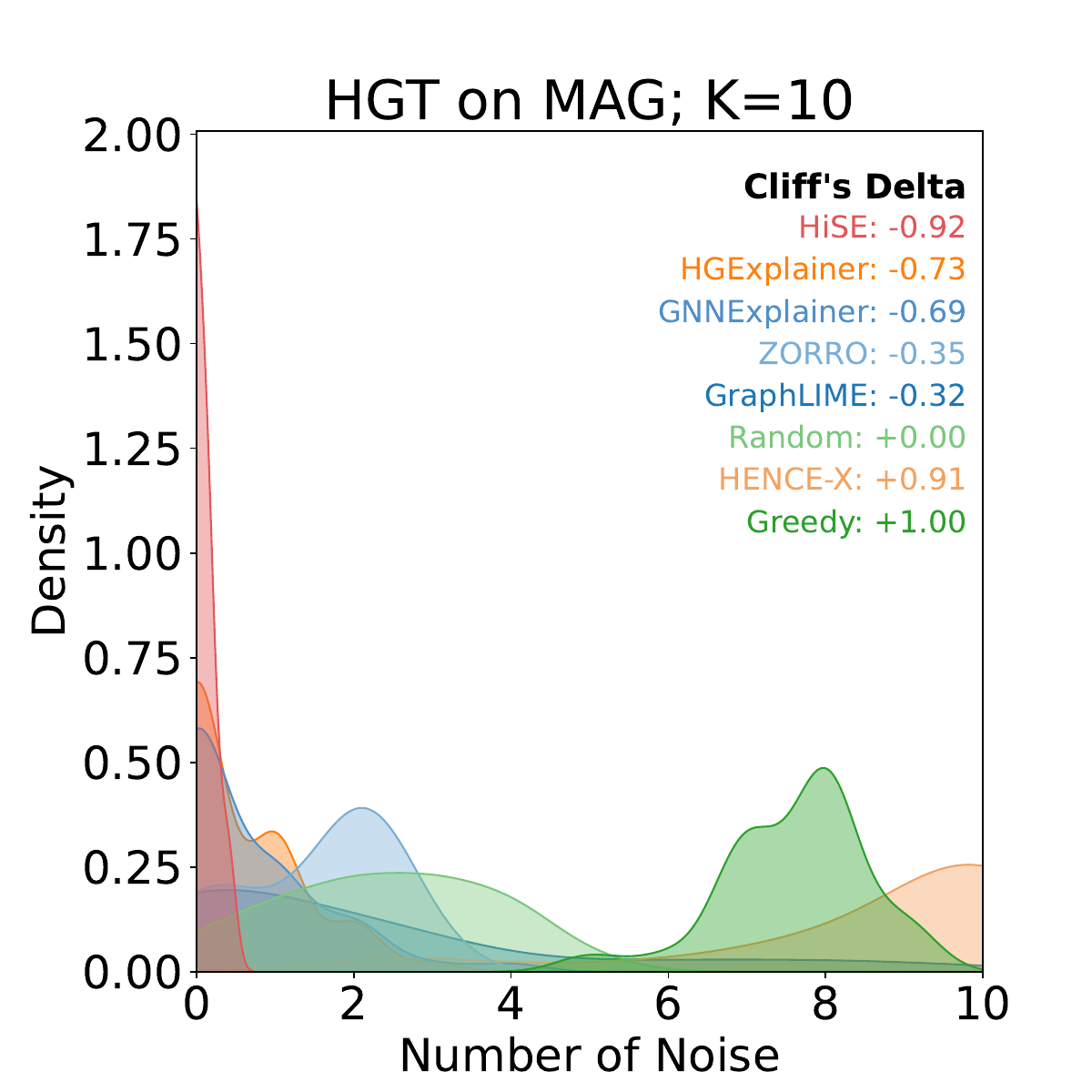}
}
\end{minipage}

\vspace{0.5em}

\begin{minipage}{\linewidth}
\centering
\includegraphics[width=1\linewidth]{pdf/Li8.pdf}
\end{minipage}

\caption{\textbf{Robustness Results}: Complete frequency distribution of noise features selected by different explainers across all HGNN models (HAN, MAGNN, HetSANN, HGT), datasets (ACM, MAG), and $K$ values ($10$, $20$). 
Each row corresponds to a different HGNN model, while the columns represent results on the ACM and MAG datasets with $K=10$ and $K=20$, respectively. 
Each subplot shows the distribution of the number of noise features appearing in the top-$K$ explanations. 
A lower concentration of noise features indicates stronger robustness of the explainer to noisy features.}
\label{fig:app_num_noise_features}
\end{figure*}

\end{document}